\documentclass{elsarticle}

\usepackage{algorithmic}
\usepackage{algorithm}
\usepackage{graphicx}
\usepackage{url}
\usepackage{multirow}
\usepackage{subfigure}
\usepackage{slashbox}
\usepackage{balance}

\usepackage{verbatim}
\usepackage{ecltree}
\usepackage{fancybox}
\usepackage{amsmath, amssymb}
\usepackage{graphicx}
\usepackage{epic}
\usepackage{listings}
\usepackage{url}
\usepackage{multirow}
\usepackage{rotating}

\newcommand{\gbf}{\boldsymbol}
\newcommand{\mbf}{\mathbf}
\newcommand{\cL}{\mathcal{L}}
\newcommand{\T}{\mathcal{T}}
\newcommand{\Z}{\mathcal{Z}}
\newcommand{\I}{\mathcal{I}}

\newcommand{\X}{\mbf{X}}
\newcommand{\A}{\gbf{\alpha}}

\newcommand{\R}{\gbf{\rho}}
\newcommand{\E}{\gbf{\eta}}
\newcommand{\N}{\gbf{\nu}}
\newcommand{\ctau}{\gbf{\tau}}
\newcommand{\cpi}{\gbf{\pi}}
\newcommand{\cphi}{\gbf{\phi}}
\newcommand{\cTheta}{\gbf{\Theta}}
\newcommand{\sumk}{\sum_{k=1}^K}
\newcommand{\sumn}{\sum_{\mu=1}^N}
\newcommand{\sumd}{\sum_{i=1}^D}
\newcommand{\sumz}{\sum_\Z}
\newcommand{\prodk}{\prod_{k=1}^K}
\newcommand{\prodn}{\prod_{\mu=1}^N}
\newcommand{\prodd}{\prod_{i=1}^D}
\newcommand{\intpi}{\int_{\cpi}}
\newcommand{\intphi}{\int_{\cphi}}
\newcommand{\eqret}{\nonumber\displaybreak[0]\\}
\newcommand{\taun}{\tau_k^\mu}
\newcommand{\xn}{x_i^\mu}
\newcommand{\etan}{\eta_{ik}}
\newcommand{\nun}{\nu_{ik}}

\newcommand{\elogrho}{\Psi(\rho_k)-\Psi(\textstyle\sum^K_{k'=1}\rho_{k'})}
 
\newcommand{\changes}[1]{}

\begin{document}

\title{Bayesian Mixture Models for Frequent Itemsets Discovery}

\author{Ruofei He}
\ead{herofff@hotmail.com}
\address{ASN Technology Group Co,.Ltd, 34 Fenghuinan Road, Xi'an, 710065, China}

\author{Jonathan Shapiro}
\ead{jls@cs.man.ac.uk}
\address{School of Computer Science, Kilburn Building, Oxford Road, University of Manchester, Manchester M13 9PL United Kingdom}

\begin{abstract}





In binary-transaction data-mining, traditional frequent itemset mining often produces results which are not straightforward to interpret. To overcome this problem, probability models are often used to produce more compact and conclusive results, albeit with some loss of accuracy. Bayesian statistics have been widely used in the development of probability models in machine learning in recent years and these methods have many advantages, including their abilities to avoid overfitting. In this paper, we develop two Bayesian mixture models with the Dirichlet distribution prior and the Dirichlet process (DP) prior to improve the previous non-Bayesian mixture model developed for transaction dataset mining. We implement the inference of both mixture models using two methods: a collapsed Gibbs sampling scheme and a variational approximation algorithm. Experiments in several benchmark problems have shown that both mixture models achieve better performance than a non-Bayesian mixture model. The variational algorithm is the faster of the two approaches while the Gibbs sampling method achieves a more accurate result. The Dirichlet process mixture model can automatically grow to a proper complexity for a better approximation. Once the model is built, it can be very fast to query and run analysis on (typically 10 times faster than Eclat, as we will show in the experiment section). However, these approaches also show that mixture models underestimate the probabilities of frequent itemsets. Consequently, these models have a higher sensitivity but a lower specificity.
\end{abstract}


\begin{keyword}
    Bayesian mixture model \sep Frequent itemsets mining \sep Gibbs sampling \sep Variational inference \sep Dirichlet process
\end{keyword}

\maketitle
\section{Introduction}
\changes{HE: Remove the sentences about finite Bayesian.}
Transaction data sets are binary data sets with rows corresponding to transactions and columns corresponding to items or attributes. Data mining techniques for such data sets have been developed for over a decade. Methods for finding correlations and regularities in transaction data can have many commercial and practical applications, including targeted marketing, recommender systems, more effective product placement, and many others.


Retail records and web site logs are two examples of transaction data sets. For example, in a retailing application, the rows of the data correspond to purchases made by various customers, and the columns correspond to different items for sale in the store. This kind of data is often sparse, i.e., there may be thousands of items for sale, but a typical transaction may contain only a handful of items, as most of the customers buy only a small fraction of the possible merchandise. In this paper we will only consider binary transaction data, but transaction data can also contain the numbers of each item purchased (multi-nomial data). An important correlation which data mining seeks to elucidate is which items co-occur in purchases and which items are mutually exclusive, and never (or rarely) co-occur in transactions. This information allows prediction of future purchases from past ones.


Frequent itemset mining and association rule mining \cite{Agrawal:1993:MAR:170036.170072} are the key approaches for finding correlations in transaction data. Frequent itemset mining finds all frequently occurring item combinations along with their frequencies in the dataset with a given minimum frequency threshold. Association rule mining uses the results of frequent itemset mining to find the dependencies between items or sets of items. If we regard the minimum frequency threshold as an importance standard, then the set of frequent itemsets contains all the ``important'' information about the correlation of the dataset. The aim of frequent itemset mining is to extract useful information from the kinds of binary datasets which are now ubiquitous in human society. It aims to help people realize and understand the various latent correlations hidden in the data and to assist people in decision making, policy adjustment and the performance of other activities which rely on correct analysis and knowledge of the data.

However, the results of such mining are difficult to use. The threshold or criterion of mining is hard to choose for a compact but representative set of itemsets. To prevent the loss of important information, the threshold is often set quite low, causing a huge set of itemsets which brings difficulties in interpretation. These properties of large scale and weak interpretability block a wider use of the mining technique and are barriers to a further understanding of the data itself. Traditionally,  Frequent Itemset Mining (FIM) suffers from three difficulties. The first is scalability, often the data sets are very large, the number of frequent item-sets of the chosen support is also large, and there may be a need to run the algorithm multiple times to find the appropriate frequency threshold. The second difficulty is that the support-confidence framework is often not able to provide the information that people really need. Therefore people seek other criteria or measurements for more ``interesting'' results. The third difficulty is in interpreting the results or getting some explanation of the data.  Therefore the recent focus of research of FIM has been in the following 3 directions.


\begin{enumerate}
    \item Looking for more compact but representative forms of the itemsets - in other words, mining compressed itemsets. The research in this direction consists of two types: lossless compression such as closed itemset mining \cite{springerlink:10.1007/3-540-49257-7-25} and lossy compression such as maximal itemset mining \cite{springerlink:10.1007/3-540-45681-3-7}. In closed itemset mining, a method is proposed to mine the set of closed itemsets which is a subset of the set of frequent itemsets. This can be used to derive the whole set of frequent itemsets without loss of information. In maximal itemset mining, the support information of the itemsets is ignored and only a few longest itemsets are used to represent the whole set of frequent itemsets.
    \item Looking for better standards and qualifications for filtering the itemsets so that the results are more ``interesting'' to users. Work in this direction focuses on how to extract the information which is both useful and unexpected as people want to find a measure that is closest to the ideal of ``interestingness''. Several objective and subjective measures are proposed such as \textit{lift} \cite{IBM1996}, $\chi^2$ \cite{Brin:1997:BMB:253262.253327} and the work of \cite{Jaroszewicz04interestingnessof} in which they use a Bayesian network as background knowledge to measure the interestingness of frequent itemsets.
    \item Looking for mathematical models which reveal and describe both the structure and the inner-relationship of the data more accurately, clearly and thoroughly. There are two ways of using probability models in FIM. The first is to build a probability model that can organize and utilize the results of mining such as the \textit{Maximal Entropy model} \cite{Tatti:2008:MEB:1453729.1453736}. The second is to build a probability model that is directly generated from the data itself which can not only predict the frequent itemsets, but also explain the data. An example of such model is the \textit{Mixture model}.
\end{enumerate}
These three directions influence each other and form the main stream of current FIM research. Of the three, the probability model solution considers the data as a sampled result from the underlying probability model and tries to explain the system in an understandable, structural and quantified way. With a good probability model, we can expect the following advantages in comparison with normal frequent itemset mining:
\begin{enumerate}
    \item The model can reveal correlations and dependencies in the dataset, whilst frequent itemsets are merely collections of facts awaiting interpretation. A probability model can handle several kinds of probability queries, such as joint, marginal and conditional probabilities, whilst frequent itemset mining and association rule mining focus only on high marginal and conditional probabilities. The prediction is made easy with a model. However, in order to predict with frequent itemsets, we still need to organize them and build a structured model first.
    \item It is easier to observe interesting dependencies between the items, both positive and negative, from the model's parameters than it is to discriminate interesting itemsets or rules from the whole set of frequent itemsets or association rules. In fact, the parameters of the probability model trained from a dataset can be seen as a collection of features of the original data. Normally, the size of a probability model is far smaller than the set of frequent itemsets. Therefore the parameters of the model are highly representative. Useful knowledge can be obtained by simply ``mining'' the parameters of the model directly.
    \item As the scale of the model is often smaller than the original data, it can sometimes serve as a proxy or a replacement for the original data. In real world applications, the original dataset may be huge and involve large time costs in querying or scanning the dataset. One may also need to run multiple queries on the data, e.g. FIM queries with different thresholds. In such circumstances, if we just want an approximate estimation, a better choice is obviously to use the model to make the inference. As we will show in this paper, when we want to predict all frequent itemsets, generating them from the model is much faster than mining them from the original dataset because the model prediction is irrelevant to the scale of the data. And because the model is independent from the minimum frequency threshold, we only need to train the model once and can do the prediction on multiple thresholds but consuming less time.
\end{enumerate}

Several probability models have been proposed to represent the data. Here we give a brief review.

The simplest and most intuitive model is the \textit{Independent model}. This assumes that the probability of an item appearing in a transaction is independent of all the other items in that transaction. The probabilities of the itemsets are products of the probabilities of the corresponding items. This model is obviously too simple to describe the correlation and association between items, but it is the starting point and base line of many more effective models.

The \textit{Multivariant Tree Distribution model} \cite{1054142}, also called the \textit{Chow-Liu Tree}, assumes that there are only pairwise dependencies between the variables, and that the dependency graph on the attributes has a tree structure. There are three steps in building the model: computing the pairwise marginals of the attributes, computing the mutual information between the attributes and applying Kruskal's algorithm \cite{1956} to find the minimum spanning tree of the full graph, whose nodes are the attributes and the weights on the edges are the mutual information between them. Given the tree, the marginal probability of an itemset can be first decomposed to a production of factors via the chains rule and then calculated with the standard belief propagation algorithm \cite{Pearl1988}.

The \textit{Maximal Entropy model} tries to find a distribution that maximizes the entropy within the constraints of frequent itemsets \cite{10.1109/TKDE.2003.1245281,Tatti:2008:MEB:1453729.1453736} or other statistics \cite{Tatti:2010:UBK:1842547.1842564}. The algorithm for solving the \textit{Maximal Entropy model} is the \textit{Iterative Scaling} algorithm. The \textit{Iterative Scaling} algorithm is a process of finding the probability of a given itemset query. The algorithm starts from an ``ignorant'' initial state and updates the parameters by enforcing them satisfying the related constraints iteratively until convergence. Finally the probability of the given query can be calculated via the parameters.

The \textit{Bernoulli Mixture model} \cite{10.1109/TKDE.2003.1245281,nla.cat-vn1670763} is based on the assumption that there are latent or unobserved types controlling the distribution of the items. Within each type, the items are independent. In other words, the items are conditionally independent given the type. This assumption is a natural extension of the \textit{Independent model}. The \textit{Bernoulli Mixture model} is a widely used model for statistical and machine learning tasks. The idea is to use an additive mixture of simple distributions to approximate a more complex distribution. This model is the focus of this paper.

When applying a mixture model to data, one needs to tune the model to the data. There are two ways to do this. In a \textit{Maximum-Likelihood Mixture Model}, which in our paper we will call the \textit{non-Bayesian Mixture Model}, the probability is characterised by a set of parameters. These are set by optimizing them to maximize the likelihood of the data. Alternatives are \textit{Bayesian Mixture models}. In these, the parameters are treated as random variables which themselves need to be described via probability distributions. Our work is focused on elucidating the benefits of Bayesian mixtures over non-Bayesian mixtures for frequent itemset mining.

Compared with non-Bayesian machine learning methods, Bayesian approaches have several valuable advantages. Firstly, Bayesian integration does not suffer from over-fitting, because it does not fit parameters directly to the data; it integrates overall parameters and is weighted by how well they fit the data. Secondly, prior knowledge can be incorporated naturally and all uncertainty is manipulated in a consistent manner. One of the most prominent recent developments in this field is the application of Dirichlet process (DP) \cite{1973} mixture model, a nonparametric Bayesian technique for mixture modelling, which allows for the automatic determination of an appropriate number of mixture components. Here, the term ``nonparametric'' means the number of mixture components can grow automatically to the necessary scale. The DP is an infinite extension of the Dirichlet distribution which is the prior distribution for finite Bayesian mixture models. Therefore the DP mixture model can contain as many components as necessary to describe an unknown distribution. By using a model with an unbounded complexity, under-fitting is mitigated, whilst the Bayesian approach of computing or approximating the full posterior over parameters mitigates over-fitting.

The difficulty of such Bayesian approaches is that finding the right model for the data is often computational intractable. A standard methodology for DP mixture model is the Monte Carlo Markov chain (MCMC) sampling. However, MCMC approach can be slow to converge and its convergence can be difficult to diagnose. An alternative is the variational inference method developed in recent years \cite{citeulike:2989074}. In this paper, we develop both finite and infinite Bayesian Bernoulli mixture models for transaction data sets with both MCMC sampling and variational inference and use them to generate frequent itemsets. We perform experiments to compare the performance of the Bayesian mixture models and the non-Bayesian mixture model. Experimental results show that Bayesian mixture model can achieve a better precision. The DP mixture model can find a proper number of mixtures automatically.

In this paper, we extend the non-Bayesian mixture model to a Bayesian mixture model. The assumption and the structure of the Bayesian model is proposed. The corresponding algorithms for inference via MCMC sampling and variational approximation are also described. For the sampling approach, we implemented Gibbs sampling algorithm \cite{Geman1984Stochastic} for the finite Bayesian mixture model (GSFBM) which is a multi-variant Markov Chain Monte Carlo (MCMC) sampling \cite{Metro1949Monte,annealing2,hastings1970montecarlo} scheme. For the variational approximation, we implement the variational EM algorithm for the finite Bayesian mixture model (VFBM) by approximating the true posterior with a factorized distribution function. We also extend the finite Bayesian mixture model to the infinite. The Dirichlet process prior is introduced to the model so that the model obtains the ability to fit a proper complexity itself. This model solves the problem of finding the proper number of components used in traditional probability models. For this model, we also implement two algorithms. The first one is Gibbs sampling for the Dirichlet Process mixture model (GSDPM). The second one is the truncated variational EM algorithm for the Dirichlet Process mixture model (VDPM). The word ``truncated'' means we approximate the model with a finite number of components.

The rest of the paper is organized as follows. In the next section, we define the problem, briefly review the development of the FIM mining and introduce the notations used in this paper. In section 3, we introduce non-Bayesian Bernoulli mixture model and its inference by EM algorithm. In section 4 and 5, we develop the Bayesian mixture models, including how to do inference via Gibbs sampling and variational EM and how to use the model for predictive inference. Then, in section 6, we use 4 benchmark transaction data sets to test the model, and compare the performances with the non-Bayesian mixture model. We also compare the MCMC approach and the EM approach by their result accuracies and time costs. Finally, we conclude this paper with a discussion of further works.

\section{Problem and Notations}\label{sec2}
\changes{HE: no changes in section 2 and 3}
Let $\I=\{i_1,i_2,\ldots,i_D\}$ be the set of items, where $D$ is the number of items. Set $I=\{i_{m_1},i_{m_2},\ldots,i_{m_k}\}\subseteq\I$ is called an \textit{itemset} with length \textit{k}, or a \textit{k-itemset}.

A transaction data set $\T$ over $\I$ is a collection of $N$ transactions: $\X^\mu\in\T, \mu=1\dots N$. A transaction $\X^\mu$ is a $D$ dimension vector:  \(x^\mu_1,\dots,x^\mu_i,\dots,x^\mu_D)\) where \(x^\mu_i\in\{0,1\}\). A transaction $\X^\mu$ is said to \textit{support} an itemset $I$ if and only if $\forall i_m\in I, x^\mu_m=1$. A transaction can also be written as an itemset. Then $\X^\mu$ \textit{supports} $I$ if $I\subseteq \X^\mu$. The frequency of an itemset is:
$$f(I)={|\{\mu|I\subseteq \X^\mu, \X^\mu\in\T\}| \over N}$$

An itemset is frequent if its frequency meets a given minimum frequency threshold: $f_{min}$. The aim of frequent itemset mining is to discover all the frequent itemsets along with their frequencies.

From a probabilistic view, the data set $\T$ could be regarded as a sampling result from an unknown distribution. Our aim is to find or approximate the probabilistic distribution which generated the data, and use this to predict all the frequent itemsets. Inference is the task of restricting the possible probability models from the data. In the Bayesian approach, this usually means putting a probability over unknown parameters. In the non-Bayesian approach, this usually means finding the best or most-likely parameters.

\section{Bernoulli Mixtures}

In this section, we describe the non-Bayesian mixture model. Consider a grocery store where the transactions are purchases of the items the store sells. The simplest model would treat each item as independent, so the probability of a sale containing item A and item B is just the product of the two probabilities separately. However, this would fail to model non-trivial correlations between the items. A more complex model assumes a mixture of independent models. The model assumes the buyers of the store can be characterized into different types representing different consumer preferences. Within each type, the probabilities are independent. In other words, the items are conditionally independent, when conditioned on the component, or type, which generated the given transaction. However, although we observe the transaction, we don't not observe the type. Thus, we must employ the machinery of inference to deal with this.

\begin{figure}[t]
    \centering
    \includegraphics[scale=0.2]{{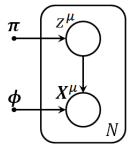}}
    \caption{non-Bayesian mixture graphic representation}\label{f1}
\end{figure}

Suppose there are $K$ components or types, then each transaction is generated by one of the $K$ components following a multinomial distribution with parameter $\gbf{\pi}=(\pi_1,\dots,\pi_K)$, where $\sum_{k=1}^K\pi_k=1$. Here we introduce a component indicator $\Z=\{z^\mu\}_{\mu=1}^N$ indicating which components the transactions are generated from: $z^\mu=k$ if $\X^\mu$ is generated from the $k$th component. According to the model assumption, once the component is selected, the probabilities of the items are independent from each other. That is, for transaction $\X^\mu$:
\begin{equation}
    p(\X^\mu|z^\mu,\cTheta)=\prod_{i=1}^D p(x^\mu_i|z^\mu,\cTheta),
\end{equation}
where $\cTheta$ representing all the parameters of the model. Thus, the probability of a transaction given by the mixture model is:
\begin{equation}\label{prob}
    p(\X^\mu|\cTheta) = \sum_{k=1}^K \pi_k\prod_{i=1}^D p(x^\mu_i|z^\mu,\cTheta)
\end{equation}
Since the transactions are binary vectors, we assume the conditional probability of each item follows a Bernoulli distribution with parameter $\phi_{ik}$:
\begin{equation}
    p(x^\mu_i|z^\mu,\cTheta)=\phi_{iz^\mu}^{x^\mu_i}(1-\phi_{iz^\mu})^{1-x^\mu_i}
\end{equation}
A graphic representation of this model is shown in Figure \ref{f1} where circles denote random variables, arrows denote dependencies, and the box (or plate) denote replication over all data points. In Figure \ref{f1}, the distribution of each transaction $\X^\mu$ depends on the selection of $z^\mu$ and model parameter $\cphi$, and $z^\mu$ depends on $\cpi$. This process will repeated $N$ times to generate the whole data set.

In this model, we need to estimate $\pi_k$ and $\phi_{ik}$ from the data. If we knew which component generated each transaction this would be easy. For example, we could estimate $\phi_{ik}$ as the frequency at which $i$ occurs in component $k$ and $\pi_k$ would be the frequency at which component $k$ occurs in the data. Unfortunately, we do not know which component generated each transaction; it is an unobserved variable. The EM algorithm \cite{1977} is often used for the parameter estimation problem for models with hidden variables in general, for mixture models in particular. We describe this in more detail in Appendix~1. For a detailed explanation, see section 9.3.3 of \cite{citeulike:873540}. The EM algorithm is given in Algorithm~\ref{em}.

\begin{algorithm}[t]
    \caption{EM algorithm for Bernoulli Mixtures}
    \label{em}
    \begin{algorithmic}
    {\small
    \STATE \textbf{initialize} $\pi_k$ and $\phi_{ik}$
    \REPEAT
        \FOR{$\mu = 1$ \text{to} $N$}
            \FOR{$k = 1$ \text{to} $K$}
                 \STATE $\taun={\pi_k\prodd\phi_{ik}^{x^\mu_i}(1-\phi_{ik})^{1-x^\mu_i} \over \sum_{k'=1}^K\pi_{k'}\prod_{i'=1}^D\phi_{i'k'}^{x^\mu_{i'}}(1-\phi_{i'k'})^{1-x^\mu_{i'}}}$
            \ENDFOR
        \ENDFOR
        \STATE $\pi_k={1 \over N}\sumn\taun$
        \STATE $\phi_{ik}={\sumn\taun\xn \over \sumn\taun}$
    \UNTIL{convergence}
    }
    \end{algorithmic}
\end{algorithm}

Another problem of this algorithm is the selection of $K$. The choice of $K$ will greatly influence the quality of the result. If the $K$ is too small, the model cannot provide accurate enough result. On the opposite, if the $K$ is too large, it may cause over-fitting problems. There is no single procedure to find out the correct $K$. People often try several increasing $K$s and determine the proper $K$ by comparing their result qualities and preventing over-fitting by cross-validation or some other criteria such as the Bayesian Information Criterion \cite{schwarz197801}.

Predicting frequent itemsets by this model is quite straightforward. For any itemset $I$, calculating its probability is done by only taking into account the items occurring in $I$  and ignoring (e.g. marginalizing over) the items which are not in $I$:
\begin{equation}
    p(I|\cTheta)=\sumk\pi_k\prod_{i_m\in I}\phi_{mk}\label{predict}
\end{equation}
The number of free parameters used for prediction is $K(D+1)-1$.

The last issue is how to generate the full set of frequent itemsets. In frequent itemset mining algorithms, obtaining the frequencies of the itemsets from the data set is always a time consuming problem. Most algorithms such as Apriori \cite{Agrawal:1994:FAM:645920.672836} require multiple scans of the data set, or use extra memory cache for maintaining special data structure such as \textit{tid\_list}s for Eclat \cite{846291} and \textit{FP-tree} for FP-growth \cite{Han:2000:MFP:342009.335372}. In the Bernoulli mixture model approach, with a prepared model, both time and memory cost can be greatly reduced with some accuracy loss since the frequency counting process has been replaced by a simple calculation of summation and multiplication.
To find the frequent itemsets using any of the probability models in this paper, simply mine the probability models instead of the data. To do this, one can use any frequent itemset datamining algorithm; we use Eclat. However, instead of measuring the frequency of the itemsets, calculate their probabilities from the probability model.

Typically this results in a great improvement in the complexity of the determination of itemset frequency.
For a given candidate itemset, to check the exact frequency of the itemset, we need to scan the original dataset for Apriori, or check the cached data structure in memory for Eclat. In both algorithms, the time complexities are $O(N)$ where $N$ is the number of transactions of the dataset. However, the calculation in mixture model merely need $KL$ times multiplication and $K$ times addition, where $L$ is the length of the itemset. Normally, $KL$ is much smaller than $N$.

The exact search strategy with Bernoulli mixture model is similar to Eclat or Apriori based on the Apriori principle \cite{Agrawal:1994:FAM:645920.672836}: \textit{All frequent itemsets' sub-itemsets are frequent, all infrequent itemsets' super-itemsets are infrequent}. Following this principle, the searching space could be significantly reduced. In our research we use the Eclat lattice decomposing framework to organize the searching process. We do not plan to discuss this framework in detail in this paper. A more specific explanation is given by \cite{846291}.

\section{The Finite Bayesian Mixtures}

\subsection{Definition of the model}
\label{sec:fbm_model}

\begin{figure}[t]
    \centering
    \includegraphics[scale=0.2]{{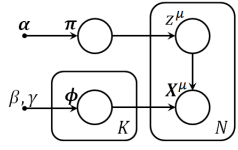}}
    \caption{finite Bayesian mixture graphic representation}\label{f2}
\end{figure}

For easier model comparison, we use the same notation in non-Bayesian
model, finite Bayesian model and the later infinite Bayesian model
when this causes no ambiguity. The difference between Bayesian mixture
models and non-Bayesian mixture models is that Bayesian mixtures try
to form a smooth distribution over the model parameters by introducing
appropriate priors. The original mixture model introduced in previous
section is a two-layer model. The top layer is the multinomial
distribution for choosing the mixtures, and the next layer is the
Bernoulli distribution for items. In Bayesian mixture we introduce a
Dirichlet distribution \cite{1973} as the prior of the multinomial
parameter $\gbf\pi$ and Beta distributions as the priors of the
Bernoulli parameters $\{\phi_{ik}\}$. The new model assumes that the
data was generated as follows.
\begin{enumerate}
    \item Assign $\alpha, \beta \text{ and } \gamma$ as the hyperparameters of the model, where $\alpha$, $\beta$ and $\gamma$ are all positive scalars. These will be chosen apriori.
    \item Choose \(\cpi\sim\)Dir(\(\alpha\)) where
        \begin{equation} \label{eq:pi|alpha}
            p(\cpi|\alpha)={\Gamma(\alpha) \over \Gamma(\alpha/K)^K}\prod^K_{k=1}\pi^{\alpha/K-1}_k
        \end{equation}
          with $\sum^K_{k=1}\pi_k=1$, $\sim$ denotes sampling, and Dir is the Dirichlet distribution.
    \item For each item and component choose \(\phi_{ik}\sim\)Beta(\(\beta,\gamma\)) where
        \begin{equation}
            p(\phi_{ik}|\beta,\gamma)={\Gamma(\beta+\gamma) \over \Gamma(\beta)\Gamma(\gamma)}\phi_{ik}^{\beta-1}(1-\phi_{ik})^{\gamma-1}
        \end{equation}
          with $\phi_{ik} \in [0,1]$ where $i\in\{1,\dots,D\}, k\in\{1,\dots,K\}$ and Beta denotes the Beta distribution.
    \item For each transaction \(\mathbf{\X}^\mu\):
    \begin{enumerate}
        \item Choose a component $z^\mu\sim$Multinomial($\cpi$), where
            \begin{equation}
                p(z^\mu=k|\cpi)=\pi_k
            \end{equation}
        \item Then we can generate data by:
            \begin{align}
                p(\X^\mu_i|z^\mu, \cphi)
                &=\prodd\phi^{x^\mu_i}_{iz^\mu}(1-\phi_{iz^\mu})^{1-x^\mu_i}
            \end{align}
    \end{enumerate}
\end{enumerate}
Figure \ref{f2} is a graphic representation for Bayesian mixtures.
 \changes{2: slight reording}

This process can be briefly written as:
\begin{align}
\cpi|\alpha&\sim \text{Dir}(\alpha/K, \alpha/K,\dots,\alpha/K)\eqret
\cphi_k|\beta,\gamma&\sim \text{Beta}(\beta,\gamma)\eqret
z^\mu|\cpi&\sim\text{Multi}(\cpi)\eqret
\X^\mu|z^\mu,\cphi&\sim p(\X^\mu|\cphi_{z^\mu})\label{model}
\end{align}
In other words, the assumption is that the data was generated by first doing the first two steps to get the parameters, then doing the second two steps $N$ times to generate the data. Since important variables of the model are not known, namely $\cpi$, $\cphi$, and $z^\mu$, the Bayesian principles say that we should compute distributions over these, and then integrate them out to get quantities of interest. However, this is not tractable. Therefore, we implement two common approximation schemes: Gibbs sampling and variational Bayes.

\subsection{Finite Bayesian mixtures via Gibbs sampling}
\label{sec:fbm_gibbs}

One approach for Bayesian inference is to approximate probabilistic integrals by sums of finite samples from the probability distribution you are trying to
Gibbs sampling is an example of the Markov chain Monte Carlo method,
which is a method of sampling from a probability. Gibbs sampling
works by sampling one component at a time. We will use a
\emph{collapsed Gibbs sampler}, which means we will not use sampling
to estimate all parameters. We will use sampling to infer the
components which generated each data point and integrate out the other
parameters.

We first introduce the inference of the model via the Gibbs sampling. Similar to the non-Bayesian mixture model, we need to work on the distribution of the component indicator $\Z$. According to the model, the joint distribution of $\Z$ is:
\begin{align}
    p(\Z)&=\int_{\cpi}p(\Z|\cpi)p(\cpi)d\cpi\eqret
         &={\Gamma(\alpha) \over \Gamma(\alpha/K)^K}\int_{\cpi}\prodk\left[\pi_k^{\alpha/K-1}\prodn\pi_k^{I(z^\mu=k)}\right]d\cpi\eqret
         &={\Gamma(\alpha) \over \Gamma(N+\alpha)}\prodk{\Gamma(N_k+\alpha/K) \over \Gamma(\alpha/K)}
\end{align}
where $N_k$ is the number of points assigned to $k$th component, the integral over $\cpi$ means the integral over a $(K-1)$-dimension simplex and the indicator function $I(z^\mu=k)$ means:
$$I(z^\mu=k)=
        \begin{cases}
        1, &\text{if } z^\mu=k\\
        0, &\text{if } z^\mu\neq k
        \end{cases}$$
The conditional probability of the $\mu$th assignment given the other assignments are:
\begin{align}
    p(z^\mu=k|\Z_{-\mu})={N_{k/\{\mu\}}+\alpha/K \over N-1+\alpha}\label{condz}
\end{align}
where $N_{k/\{\mu\}}$ is the number of points assigned to $k$th component except the $\mu$th point. The posterior distribution of the Bernoulli parameter $\cphi_k$ is the following if we know the component assignment:
\begin{align}
    p(\cphi_k|\Z, \T)&\propto p(\T|\cphi_k,\Z)p(\cphi_k|\beta,\gamma)\\
                    &\propto \prodd \text{Beta}(\phi_{ik}|\beta_{ik},\gamma_{ik})\label{postphi}
\end{align}
where
\begin{align}
    &\beta_{ik}=\beta+\sum_{\mu=1}^NI(z^\mu=k)x_i^\mu\eqret
    &\gamma_{ik}=\gamma+N_k-\sum_{\mu=1}^NI(z^\mu=k)x_i^\mu\nonumber
\end{align}
Combining Equation (\ref{condz}) and (\ref{postphi}), we can calculate the posterior probability of the $\mu$th assignment by integrating out $\cphi$:
\begin{align}
   &p(z^\mu=k|\Z_{-\mu},\T)=\int_{\cphi_k} p(z^\mu=k|\Z_{-\mu})p(\cphi_k|\Z_{-\mu}, \T)d\cphi_k\eqret
                    &\propto{N_{k/\{\mu\}}+\alpha/K \over N-1+\alpha}\prodd\left({\beta_{ik/\{\mu\}}\over \beta+\gamma+N_k}\right)^{x^\mu_i}\left({\gamma_{ik/\{\mu\}}\over \beta+\gamma+N_k}\right)^{1-x^\mu_i}\label{finGibbs}
\end{align}
where $N_{k/\{\mu\}}$,$\beta_{ik/\{\mu\}}$ and $\gamma_{ik/\{\mu\}}$
are calculated excluding the $\mu$th point and the integral over
$\cphi_k$ means integral over a $D$-dimension vector
$\cphi_k\in[0,1]^D$. Equation~(\ref{finGibbs}) shows how to sample the
component indicator based on the other assignments of the
transactions.The whole process of the collapsed Gibbs sampling for the
finite Bayesian mixture model is shown in Algorithm
\ref{fgs}. Initialization of parameters $\alpha, \beta,$ and $\gamma$
is discussed in section~\ref{sec:empir-results-disc}.
\begin{algorithm}[t]
    \caption{collapsed Gibbs sampling for finite Bayesian mixture model}
    \label{fgs}
    \begin{algorithmic}
    {\small
    \STATE \textbf{input parameters}  $\alpha$, $\beta$, $\gamma$
    \STATE \textbf{input parameter} $K$ as the number of components
    \STATE \textbf{initialize} $\Z$ to be a random assignment
    \REPEAT
        \FOR{$\mu = 1$ \text{to} $N$}
            \STATE For all $i,k$ update $\beta_{ik}, \gamma_{ik}$ by
            \STATE $\quad\beta_{ik}=\beta+\sum_{\mu'\neq\mu}^NI(z^{\mu'}=k)x_i^{\mu'}$
            \STATE $\quad\gamma_{ik}=\gamma+n_{k/\{\mu\}}-\sum_{\mu'\neq\mu}^NI(z^{\mu'}=k)x_i^{\mu'}$
            \STATE For all $k$ calculate multinomial probabilities based on
            \STATE $\quad p(z^\mu=k)=\frac{\alpha/K+n_{k/\{\mu\}}}{\alpha+N-1}\prodd\left({\beta_{ik}\over \beta+\gamma+N_k}\right)^{x^\mu_i}\left({\gamma_{ik}\over \beta+\gamma+N_k}\right)^{1-x^\mu_i}$
            \STATE Normalize $p(z^\mu=k)$ over $k$
            \STATE Sample $z^\mu$ based on $p(z^\mu)$
        \ENDFOR
    \UNTIL{convergence}
    }
    \end{algorithmic}
\end{algorithm}

The predictive inference after Gibbs sampling is quite straightforward. We can estimate the proportion and the conditional probability parameters by the sampling results. The proportion is inferred from the component indicator $\Z$ we sampled:
\begin{equation}
    \pi_k=\frac{N_k+\alpha/K}{N+\alpha}
\end{equation}
The conditional Bernoulli parameters are estimated as following:
\begin{equation}
    \phi_{ik}=\frac{\beta+\sum_{\mu=1}^NI(z^\mu=k)x_i^\mu}{\beta+\gamma+N_k}
\end{equation}
For a given itemset $I$, its predictive probability is:
\begin{equation}
    p(I|\Z)=\sum_{k=1}^{K}\pi_k\prod_{i_m\in I}\phi_{i_mk}
\end{equation}
In practice, the parameters $\pi_k$ and $\phi_{ik}$ only need to be
calculated only once for prediction. The model contains $K\times
(D+1)-1$ free parameters.

\subsection{Finite Bayesian Mixture Model via Variational Inference}
\label{sec:finite-bayes-mixt}
In this section we describe the variational EM algorithm \cite{Beal2003Variational,citeulike:873540} for this model. Based on the model assumption, the joint probability of the transaction $\X^\mu$, components indicator $z^\mu$ and the model parameters $\gbf\pi$ and $\gbf\phi$ is:
\begin{align}
    p(\mbf{X}^\mu,z^\mu,\cpi,\cphi|\alpha,\beta,\gamma)
    =p(\mbf{X}^\mu|z^\mu,\cphi)p(z^\mu|\cpi)p(\cphi|\beta,\gamma)p(\cpi|\alpha)
\end{align}
For the whole data set:
\begin{align}\label{eq:jointVB}
    p(\T,\Z,\cpi,\cphi|\alpha,\beta,\gamma)
    =\prodn[p(\mbf{X}^\mu|z^\mu,\cphi)p(z^\mu|\cpi)]p(\cphi|\beta,\gamma)p(\cpi|\alpha)
\end{align}
Integrating over $\cpi$, $\cphi$, summing over $\mathcal{Z}$ and
taking the logarithm, we obtain the log-likelihood of the data set:
\begin{align} \label{eq:integralEM}
    \ln p(\T|\alpha,\beta,\gamma)
    =\ln\int_{\cpi}\int_{\cphi}\sum_{\Z}p(\T,\Z,\cpi,\cphi|\alpha,\beta,\gamma)d\cphi d\cpi
\end{align}
Here the integral over $\cpi$ means integral over a $(K-1)$-dimension
simplex. The integral over $\cphi$ means integral over a $K\times D$
vector $\cphi\in[0,1]^{K\times D}$. The summing over $\mathcal{Z}$ is
summing over all possible $\mathcal{Z}$ configurations. This integral
is intractable because of the coupling of $\Z$ and $\cpi$. This approximate distribution is chosen so that: the
variables are decoupled, and the approximate distribution is a close
as possible to the true distribution. In other words, the task is to
find the decoupled distribution most like the true distribution, and
use the approximate distribution to do inference.

We assume the distribution has the following form:
\begin{align}\label{eq:q}
    &q(\mathcal{Z},\cpi,\cphi|\ctau,\R,\E,\N)\nonumber\\
    &\qquad=\left[\prodn q(z^\mu|\ctau^\mu)\right]\cdot\left[\prodd\prodk q(\phi_{ik}|\eta_{ik},\nu_{ik})\right]q(\cpi|\R)\\
&\text{where}\nonumber\\
&\qquad q(z^\mu|\ctau^\mu)\sim\text{Multinomial}(\ctau^\mu)\nonumber\\
&\qquad q(\phi_{ik}|\eta_{ik},\nu_{ik})\sim\text{Beta}(\eta_{ik},\nu_{ik})\nonumber\\
&\qquad q(\cpi|\R)\sim\text{Dir}(\R)\nonumber
\end{align}
Here $\R$, $\E$ and $\N$ are free variational parameters corresponding to the hyperparameters $\alpha$, $\beta$ and $\gamma$, and $\ctau$ is the multinomial parameter for decoupling $\cpi$ and $\Z$. We use this $q(\cdot)$ function to approximate the true posterior distribution of the parameters. To achieve this, we need to estimate the values of $\R$, $\E$ and $\N$. Similar to non-Bayesian mixture EM, we expand the log-likelihood and optimize its lower bound.
The optimization process is quite similar to the calculations we did in non-Bayesian EM part. In the optimization, we use the fact that $E[\log\pi_k]=\Psi(\alpha_k)-\Psi(\textstyle\sum_{k'=1}^K\alpha_{k'})$ if $\cpi\sim$ Dir($\A$) where $\Psi(\cdot)$ is the digamma function. This yields:
\begin{align}
    \rho_k=&\alpha+\textstyle{\sumn}\taun\label{rhok}\\
    \etan=&\beta+\textstyle{\sumn}\taun\xn\label{etan}\\
    \nun=&\gamma+\textstyle{\sumn}\taun(1-\xn)\label{nun}\\
    \taun\propto&\exp\left\{\elogrho\right.\eqret &\left.+\textstyle{\sumd} x_i^\mu[\Psi(\eta_{ik})-\Psi(\eta_{ik}+\nu_{ik})]\right.\nonumber\\
    &\left.+\textstyle{\sumd}(1-x_i^\mu)[\Psi(\nu_{ik})-\Psi(\eta_{ik}+\nu_{ik})]\right\}\label{Sk}
\end{align}
Equation (\ref{rhok}) to (\ref{Sk}) form an iterated optimization procedure. A brief demonstration of this procedure is given by Algorithm \ref{VEMF2}.
\begin{algorithm}[t]
    \caption{Variational EM for Finite Bayesian Bernoulli Mixtures}\label{VEMF2}
    \begin{algorithmic}
    \STATE \textbf{input parameters} $\alpha$, $\beta$ and $\gamma$
    \STATE \textbf{input parameters} $K$ as the number of components
    \STATE \textbf{initialize} $\taun$ to be a random assignment
    \REPEAT
        \STATE For all $i,k$ update $\rho_{k}, \etan, \nun$ by
        \STATE $\rho_k=\alpha+\textstyle{\sumn}\taun$
        \STATE $\etan=\beta+\textstyle{\sumn}\taun\xn$
        \STATE $\nun=\gamma+\textstyle{\sumn}\taun(1-\xn)$
        \FOR{$\mu = 1$ \text{to} $N$}
            \FOR{$k = 1$ \text{to} $K$}
                 \STATE Update $\tau_k^{\mu}$ according to (\ref{Sk})
            \ENDFOR
            \STATE Normalize $\taun$ over $k$
        \ENDFOR
    \UNTIL{convergence}
    \end{algorithmic}
\end{algorithm}

For any itemset $I$, its predictive probability given by the model is:
\begin{align}
    p(I|q)=&\intpi\intphi\sum_{\hat{z}} p(I|\hat{z},\cphi)p(\hat{z}|\cpi)q(\cpi,\cphi|\R,\E,\N)d\cpi d\cphi\nonumber\\
    =&\sumk\frac{\rho_k}{\sumk \rho_{k'}}\prod_{i_m\in I}\frac{\eta_{mk}}{\eta_{mk}+\nu_{mk}}\label{BayesianPredict}
\end{align}
In Equation (\ref{BayesianPredict}), we use the decoupled $q(\cdot)$ to replace the true posterior distribution so that the integral is solvable. Equation (\ref{BayesianPredict}) shows that when doing predictive inference, we only need to take care the value of $\rho_k$, $\etan$ and $\nun$ proportionally. Therefore the number of parameters is exactly the same as the non-Bayesian model.

\section{The Dirichlet Process Mixture Model}
\label{sec:dirichl-proc-mixt}
\changes{HE: Complete new section. I first explain the difference from non-Bayesian to Bayesian model, then develop the inference under finite circumstance. Based on the finite sampling scheme, I extend K to infinity and obtain the DP sampling scheme in an intuitive way. At last, I briefly explain DP, mentioned the urn scheme and rewrite the DP mixture model.}
The finite Bayesian mixture model is still restricted by the fact that
the number of components $K$ must be chosen in advance. Ferguson
\cite{1973} proposed the Dirichlet Process (DP) as the infinite
extension of the Dirichlet distribution. Applying the DP as the prior
of the mixture model allows us to have an arbitrary number of
components,  growing as necessary during the learning process. In the finite Bayesian mixture model, the Dirichlet distribution is a prior for choosing components. Here the components are in fact distributions drawn from a base distribution Beta($\beta$, $\gamma$). In Dirichlet distribution, the number of components is a fixed number $K$. So each time we draw a distribution, the result is equal to one of the $K$ distributions drawn from the base distribution with probabilities given by the Dirichlet distribution. Now we relax the number of components as unlimited and keep the discreteness of the components, which means that each time we draw a distribution (component), the result is either equal to an existed distribution or a new draw from the base distribution. This new process is called the Dirichlet Process \cite{1973} and the drawing scheme is the Blackwell-MacQueen's P\'{o}lya urn scheme \cite{blackwell:polya}:
\begin{align}
z^\mu=
    \begin{cases}
        k\quad \text{with prob. } {N_{k/\{\mu\}} \over N-1+\alpha}\\
        K+1,\;\;\cphi_{K+1}\sim \text{Beta}(\beta,\gamma)\quad\text{with prob. } {\alpha\over N-1+\alpha}
    \end{cases}\label{polya}
\end{align}
The previous model should also be rewritten as:
\begin{align}
B|\alpha, B_0&\sim \text{DP}(\alpha, B_0(\beta,\gamma))\eqret
\cphi^\mu|B&\sim B\eqret
\X^\mu|\cphi^\mu&\sim p(X^\mu|\cphi^\mu)\label{DPmodel}
\end{align}

\subsection{The Dirichlet Process Mixture Model via Gibbs Sampling}
Based on the P\'{o}lya urn scheme we can allow $K$ to grow. Following this scheme, every time we draw a distribution, there is a chance that the distribution comes from the base distribution, therefore adding a new component to the model. This scheme makes the $K$ has the potential to grow to any positive integer.

Assume at a certain stage, the actual number of components is $K$. Based on Equation (\ref{polya}):
\begin{align}
    p(z^\mu=k|\Z_{-\mu})={N_{k/\{\mu\}} \over N-1+\alpha},\text{iff }k\leq K\nonumber
\end{align}
Then the probability that the $\mu$th point is in a new component is:
\begin{align}
    p(z^\mu=K+1|\Z_{-\mu})=1-p(z^\mu\leq K|\Z_{-\mu})={\alpha \over N-1+\alpha}\nonumber
\end{align}
The rest of the posterior probability remains the same, as there is no $K$ involved:
\begin{align}
   &p(z^\mu=k|\Z_{-\mu},\T)\eqret
                    &\propto{N_{k/\{\mu\}} \over N-1+\alpha}\prodd\left({\beta_{ik}\over \beta+\gamma+N_k}\right)^{x^\mu_i}\left({\gamma_{ik}\over \beta+\gamma+N_k}\right)^{1-x^\mu_i}\label{infinGibbs}
\end{align}
For the new component, $N_{K+1}=0$ and we have,
\begin{align}
    &p(z^\mu=K+1|\Z_{-\mu},\T)\eqret&\propto{\alpha\over N-1+\alpha}\prodd\left({\beta\over \beta+\gamma}\right)^{x^\mu_i}\left({\gamma\over \beta+\gamma}\right)^{1-x^\mu_i}\label{newcom}
\end{align}
Equation (\ref{infinGibbs}) and (\ref{newcom}) form a collapsed Gibbs sampling scheme. At the beginning, all data points are assigned to one initial component. Then for each data point in the data set, the component indicator is sampled according to the posterior distribution provided by Equation (\ref{infinGibbs}) and (\ref{newcom}). After the indicator is sampled, the relevant parameters $N_k$, $\beta_{ik}$ and $\gamma_{ik}$ are updated for next data point. The whole process will keep running until some convergence condition is met. Algorithm \ref{igs} describes the method.

\begin{algorithm}[t]
    \caption{collapsed Gibbs sampling for Dirichlet process mixture model}
    \label{igs}
    \begin{algorithmic}
    {\small
    \STATE \textbf{input parameters} $\alpha$, $\beta$, $\gamma$
    \STATE \textbf{initialize} $K=1$
    \REPEAT
        \FOR{$\mu = 1$ \text{to} $N$}
            \STATE For all $i, k \text{ with } 1\leq k\leq K$, update $\beta_{ik}, \gamma_{ik}$ by
            \STATE $\quad\beta_{ik}=\beta+\sum_{\mu'\neq\mu}^N\delta(z^{\mu'}-k)x_i^{\mu'}$
            \STATE $\quad\gamma_{ik}=\gamma+n_{k/\{\mu\}}-\sum_{\mu'\neq\mu}^N\delta(z^{\mu'}-k)x_i^{\mu'}$
            \STATE Calculate multinomial probabilities based on
            \STATE $
       p(z^\mu=k|\Z_{-\mu},\T)\propto
        \begin{cases}
            \frac{n_{k/\{\mu\}}}{\alpha+N-1}\prodd\left({\beta_{ik}\over \beta+\gamma+N_k}\right)^{x^\mu_i}\left({\gamma_{ik}\over \beta+\gamma+N_k}\right)^{1-x^\mu_i}, &\text{if } z^\mu=K, k\leq K\\
            \frac{\alpha}{\alpha+N-1}\prodd\left({\beta \over \beta+\gamma}\right)^{x^\mu_i}\left({\gamma\over \beta+\gamma}\right)^{1-x^\mu_i}, &\text{if }  z^\mu=K+1
        \end{cases}
$

            \STATE Normalize $p(z^\mu=k)$ over $K+1$
            \STATE Sample $z^\mu$ based on $p(z^\mu)$
            \IF{component $K+1$ selected}  \STATE $K=K+1$ \ENDIF
        \ENDFOR
    \UNTIL{convergence}
    }
    \end{algorithmic}
\end{algorithm}

The predictive inference is generally the same as the finite version.

\subsection{DP Mixtures via Variational Inference}
\label{sec:dp-mixtures-var}
\changes{HE: As the variational inference for finite Bayesian is removed, I add some details about the calculation}
 Although the Gibbs sampler can provide a very accurate approximation
 to the posterior distribution for the component indicators, it needs
 to update the relative parameters for every data point. Thus it is
 computational expensive and not very suitable for large scale
 problems. In 1994, Sethuraman developed the stick-breaking
 representation \cite{sethuraman:stick} of DP which captures the DP
 prior most explicitly among other representations. In the
 stick-breaking representation, an unknown random distribution is
 represented as a sum of countably infinite atomic distributions. The
 stick-breaking representation provide a possible way for doing the
 inference of DP mixtures by variational methods. A variational method
 for DP mixture has been proposed by
 \cite{Blei05variationalinference}. They showed that the variational
 method produced comparable result to MCMC sampling algorithms including the collapsed
 Gibbs sampling, but is much faster.

\begin{figure}[t]
    \centering
    \includegraphics[scale=0.2]{{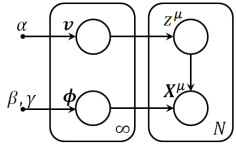}}
    \caption{Graphic representation of DP mixture in stick-breaking representation}\label{f3}
\end{figure}

 In the transaction data set background, the target distribution is the distribution of the transaction $p(\X^\mu)$ and the atomic distributions are the conditional distributions such as $p(\X^\mu|z^\mu)$. Based on the stick-breaking representation, the Dirichlet process mixture model is the following.
\begin{enumerate}
    \item Assign $\alpha$ as the hyperparameter of the Dirichlet process, $\beta$, $\gamma$ as the hyperparameters of the base Beta distribution, where they are all positive scalars.
    \item Choose $v_k\sim \text{Beta}(1,\alpha),k=1,...,\infty$
    \item Choose $\phi_{ik}\sim \text{Beta}(\beta,\gamma),i=1,\dots,D;k=1,\dots$
    \item For each transaction \(\mathbf{\X}^\mu\):
        \begin{enumerate}
        \item Choose a component $z^\mu\sim$Multinomial($\cpi(\mathbf{v})$) where
            \begin{equation}
                \pi_k(\mathbf{v})=v_k\prod_{l=1}^{k-1}(1-v_l)
            \end{equation}
        \item Then we can generate data by:
            \begin{align}
                p(\X^\mu|z^\mu, \cphi)
                &=\prodd\phi^{x^\mu_i}_{iz^\mu}(1-\phi_{iz^\mu})^{1-x^\mu_i}
            \end{align}
        \end{enumerate}
\end{enumerate}
The stick-breaking construction for the DP mixture is depicted in Figure \ref{f3}.
With the model assumption, the joint probability of the data set $\T$, components indicators $\Z$ and the model parameters $\mathbf{v}$ and $\gbf\phi$ is:
\begin{align}\label{eq:jointDP}
    &p(\T,\Z,\mathbf{v},\cphi|\alpha,\beta,\gamma)\nonumber\\
    &\quad=\prodn[p(\X^\mu|z^\mu,\cphi)p(z^\mu|\mathbf{v})]p(\cphi|\beta,\gamma)p(\mathbf{v}|\alpha)
\end{align}
Integrating over $\cpi$, $\cphi$, summing over $\mathcal{Z}$ and
applying the logarithm, we obtain the log-likelihood of the data set:
\begin{align} \label{eq:integralDP}
    \ln p(\T|\alpha,\beta,\gamma)
    =\ln\int_{\mathbf{v}}\int_{\cphi}\sum_{\Z}p(\T,\Z,\mathbf{v},\cphi|\alpha,\beta,\gamma)d\cphi d\mathbf{v}
\end{align}
Here the integral over $\mathbf{v}$ means integral over a vector $\mathbf{v}\in[0,1]^\infty$. The integral over $\cphi$ means integral over a $\infty\times D$ vector $\cphi\in[0,1]^{\infty\times D}$. The summing over $\mathcal{Z}$ is summing over all possible $\mathcal{Z}$ configurations. This integral is intractable because of the integral over infinity dimensions and the coupling of $\Z$ and $\mathbf{v}$.

Notice the following limit with a given truncation $K$:
\begin{align}
\lim_{K\rightarrow\infty}[1-\sum_{k=1}^K\pi_k(\mbf{v})]=\lim_{K\rightarrow\infty}\prod_{k=1}^K(1-v_k)=0\label{lim}
\end{align}
Equation (\ref{lim}) shows that for a large enough truncation level $K$, all the components beyond the $K$th component could be ignored as the sum of their proportion is very close to 0, which means that it is possible to approximate the infinite situation by a finite number of components. The difference with finite Bayesian model is that in finite Bayesian mixture, the number of component is finite; but in truncated DP mixture, the number of component is infinite. We only use a finite distribution to approximate it. Therefore we can use a finite and fully decoupled function as the approximation of true posterior distribution. We propose the following factorized family of variational distribution:
\begin{align}
    &q(\mathcal{Z},\mbf{v},\cphi|\ctau,\R_1,\R_2,\E,\N)\eqret
    &\quad=\left[\prodn q(z^\mu|\ctau^\mu)\right]\left[\prod_{k=1}^K\prodd q(\phi_{ik}|\eta_{ik},\nu_{ik})\right]\prod_{k=1}^{K-1}q(v_k|\rho_{1k},\rho_{2k})\label{qfun}\\
    &\text{where}\eqret
    &\quad q(z^\mu|\ctau^\mu)\sim\text{Multinomial}(\ctau^\mu)\eqret
    &\quad q(\phi_{ik}|\eta_{ik},\nu_{ik})\sim\text{Beta}(\eta_{ik},\nu_{ik})\eqret
    &\quad q(v_k|\rho_{1k},\rho_{2k})\sim\text{Beta}(\rho_{1k},\rho_{2k})\nonumber
\end{align}
Here $\R_1$, $\R_2$, $\E$ and $\N$ are free variational parameters corresponding to the hyperparameters 1, $\alpha$, $\beta$ and $\gamma$, and $\ctau$ is the multinomial parameter for decoupling $\mbf{v}$ and $\Z$. As we are assuming the proportion of the components beyond $K$ is 0, the value of $v_K$ in the approximation is always 1. We use this $q(\cdot)$ function to approximate the true posterior distribution of the parameters. To achieve this, we need to estimate the values of $\R_1$, $\R_2$, $\E$ and $\N$.
A detailed computation of the optimization is given by \cite{Blei05variationalinference}.
The optimization yields:
\begin{align}
    \rho_{1k}=&1+\textstyle{\sumn}\taun\label{rho1k}\\
    \rho_{2k}=&\alpha+\textstyle{\sumn\sum_{k'=k+1}^K}\tau_{k'}^\mu\label{rho2k}\\
    \etan=&\beta+\textstyle{\sumn}\taun\xn\label{etanDP}\\
    \nun=&\gamma+\textstyle{\sumn}\taun(1-\xn)\label{nunDP}\\
    \taun\propto&\exp\left\{\Psi(\rho_{1k})+\textstyle{\sum_{k'=1}^{k-1}\Psi(\rho_{2k'})-\textstyle{\sum_{k'=1}^{k}}\Psi(\rho_{1k'}+\rho_{2k'})}\right.\nonumber\\
    &\left.+\textstyle{\sumd} x_i^\mu[\Psi(\eta_{ik})-\Psi(\eta_{ik}+\nu_{ik})]\right.\nonumber\\
    &\left.+\textstyle{\sumd}(1-x_i^\mu)[\Psi(\nu_{ik})-\Psi(\eta_{ik}+\nu_{ik})]\right\}\label{taudp}
\end{align}
Equation (\ref{rho1k}) to (\ref{taudp}) form an iterated optimization procedure. A brief demonstration of this procedure is given by Algorithm \ref{VDP2}.
\begin{algorithm}[t]
    \caption{Variational EM for DP Bernoulli Mixtures}\label{VDP2}
    \begin{algorithmic}
    \STATE \textbf{input parameters} $\alpha$, $\beta$ and $\gamma$
    \STATE \textbf{input parameter} $K$ as the truncated number of components
    \STATE \textbf{initialize} $\taun$ to be a random assignment
    \REPEAT
        \STATE For all $i,k$ update $\rho_{1k}, \rho_{2k}, \etan, \nun$ by
        \STATE $\rho_{1k}=1+\textstyle{\sumn}\taun$
        \STATE $\rho_{2k}=\alpha+\textstyle{\sumn\sum_{k'=k+1}^K}\tau_{k'}^\mu$
        \STATE $\etan=\beta+\textstyle{\sumn}\taun\xn$
        \STATE $\nun=\gamma+\textstyle{\sumn}\taun(1-\xn)$
        \FOR{$\mu = 1$ \text{to} $N$}
            \FOR{$k = 1$ \text{to} $K$}
                 \STATE Update $\tau_k^{\mu}$ according to (\ref{taudp})
            \ENDFOR
            \STATE Normalize $\taun$ over $k$
        \ENDFOR
    \UNTIL{convergence}
    \end{algorithmic}
\end{algorithm}

The predictive inference is given by Equation (\ref{DPPredict}). Same
as we did in finite model, we use the decoupled $q(\cdot)$ function to
replace the true posterior distribution so that we can do the integral
analytically. In fact we only need to use the value of
$\frac{\rho_{1k}}{\rho_{1k}+\rho_{2k}}\prod_{k'=1}^{k-1}\frac{\rho_{2k'}}{\rho_{1k'}+\rho_{2k'}}$
as the proportion of each component. Thus the number of parameters
used for prediction is still the same as the finite model if we set
the truncation level to be the same value as the number of components $K$ in the finite model.
\begin{align}
    p(I|q)=&\int_{\mbf{v}}\intphi\sum_{\hat{z}} p(I|\hat{z},\cphi)p(\hat{z}|\mbf{v})q(\mbf{v},\cphi|\R_1,\R_2,\E,\N)d\mbf{v} d\cphi\nonumber\\
    =&\sum_{k=1}^T\frac{\rho_{1k}}{\rho_{1k}+\rho_{2k}}\prod_{k'=1}^{k-1}\frac{\rho_{2k'}}{\rho_{1k'}+\rho_{2k'}}\prod_{i_m\in I}\frac{\eta_{mk}}{\eta_{mk}+\nu_{mk}}\label{DPPredict}
\end{align}
\section{Empirical Results and Discussion}
\label{sec:empir-results-disc}
In this section, we compare the performances of proposed models with the non-Bayesian mixture model using 5 synthetic data sets and 4 real benchmark data sets. We generate five synthetic datasets from five mixture models with 15, 25, 50, 75 and 140 components respectively and apply the four methods to the synthetic datasets to see how closely the new models compare with the original mixture model. For the real data sets, we choose the mushroom, chess, Anonymous Microsoft Web data \cite{Frank+Asuncion:2010} and accidents \cite{geurts03using}. The data sets mushroom and chess we used were transformed to discrete binary data sets by Roberto Bayardo and the transformed version can be downloaded at \url{http://fimi.ua.ac.be/data/}. In Table \ref{tab_dat} we summarize the main characteristics of these 4 data sets.
\begin{table}[t]
        \centering
    \begin{tabular}{|c|c|c|c|c|}
        \hline
        Name & $N$ & $D$ & $N_{1's}$ & $Density$ \\
        \hline
        chess & 3197 & 75 & 118252 & 49.32\%\\
        mushroom & 8125 & 119 & 186852 & 19.33\%\\
        MS Web Data & 37711 & 294 & 113845 & 1.03\%\\
        accidents & 341084 & 468 & 11500870 & 7.22\%\\
        \hline
    \end{tabular}
    \caption{General Characteristics of the testing data sets: $N$ is the number of records, $D$ is the number of items, $N_{1's}$ is the number of ``1"s and the $Density$ reflects the sparseness of the data set which is calculated by $Density=N_{1's}/(ND)$}\label{tab_dat}

\end{table}
We randomly sampled a proportion of the data sets for training and used the rest for testing. For synthetic data sets, mushroom and chess, we sampled half of the data and used the rest for testing. For MS Web data, the training and testing data sets were already prepared as 32711 records for training and 5000 records for testing. For accidents, as this data set is too large to fit into the memory, we sampled 1/20 as training data and sampled another 1/20 for testing. We use the following 3 evaluation criteria for model comparison.
\begin{enumerate}
    \item We measure the difference between the predicted set of frequent itemsets and the true set of frequent itemsets by calculating the false negative rate ($F^-$) and the false positive rate ($F^+$). They are calculated by
$$F^-=\frac{N_M}{N_M+N_C}, F^+=\frac{N_F}{N_F+N_C}$$ where $N_M$ is
the number of itemsets that the model failed to predict, $N_F$ is the
number of itemsets that the model falsely predicted and $N_C$ is the
number of itemsets that the model predicted correctly. Note that
$1-F^{-}$ gives the \emph{recall} and $1-F^{=}$ gives the
\emph{precision} of the data-miner.
    \item For any true frequent itemset $I$, we calculate the relative error by:
            $$e(I)={|p_M(I)-f(I)| \over f(I)},$$
        where $p_M(I)$ is the probability predicted by the model.
        The overall quality of the estimation $\hat{E}$ is:
        \begin{equation}
            \hat{E}={1 \over N_I}\sum_{j=1}^{N_I}e(I_j),\label{crit}
        \end{equation}
        where $N_I$ is the total number of true frequent itemsets.
    \item To test whether the model is under-estimating or over-estimating, we define the empirical mean of relative difference for a given set $S$ as:
        \begin{equation}
            \hat{D}_S={1 \over |S|}\sum_{j=1}^{|S|}{p_M(I_j)-f(I_j) \over f(I_j)}\label{crit2}
        \end{equation}
\end{enumerate}

The parameter settings of the experiments are as follows. As the aim of applying the algorithms on the synthetic datasets is to see how closely the new models compare with the original mixture model, we assume that we already know the correct model before learning. Therefore for the synthetic data sets, we choose the number of components the same as the original mixture model except the DP mixture via Gibbs sampling. For the real datasets, we used 15, 25, 50 and 75 components respectively for the finite Bayesian models and the truncated DP mixture model. For the DP mixture model via Gibbs sampler, we don't need to set $K$. For each parameter configuration, we repeat 5 times to reduce the variance. The hyper-parameters for both finite and infinite Bayesian models are set as follows: $\alpha$ equals 1.5, $\beta$ equals the frequency of the items in the whole data sets and $\gamma$ equals $1-\beta$.
\begin{table}[t]
    \centering
\begin{tabular}{|c|cccc|}
\hline
    & chess & mushroom & MS Web & accidents\\
\hline
threshold & 50\% & 20\% & 0.5\% & 30\%\\
\hline
\backslashbox{length\kern-1em}{\kern-1em total} & 1262028 & 53575 & 570 & 146904\\
\hline
1 & 37 & 42 & 79 & 32\\
2 & 530 & 369 & 214 & 406\\
3 & 3977 & 1453 & 181 & 2545\\
4 & 18360 & 3534 & 85 & 9234\\
5 & 57231 & 6261 & 11 & 21437\\
6 & 127351 & 8821 & 0 & 33645\\
7 & 209743 & 10171 & 0 & 36309\\
8 & 261451 & 9497 & 0 & 26582\\
9 & 249427 & 7012 & 0 & 12633\\
10 & 181832 & 4004 & 0 & 3566\\
$>$10 & 152089 & 2411 & 0 & 515\\
\hline
\end{tabular}
\caption{Minimum frequency threshold and the number of frequent itemsets}\label{fi} 
\end{table}

The last parameter for the experiments is the minimum frequency threshold. As we mentioned, in practical situation, there is no standard procedure to select this threshold. However in our experiments, as the goal is to test our models, the requirement of the threshold is that we need to make the itemsets frequent enough to represent the correlation within the data sets, while generating enough number of frequent itemsets for model comparison as well. The threshold also should not be too low as a low threshold may make the experiments taking too much time. According to the characteristics of the datasets and several test runs, we set the thresholds of the data sets as in Table \ref{fi} so that the numbers of the frequent itemsets are proper for model evaluation. The numbers of the frequent itemsets with different lengths are also listed. For the synthetic datasets, the minimal support threshold are 30\%.

The test results of the synthetic datasets are shown in Table \ref{resultSyn15} where $F^-$ is the \textit{False Negative Rate} in percentage, $F^+$ is the \textit{False Positive Rate} in percentage and the $\hat{E}$ is the \textit{Empirical Relative Error} in percentage. We also calculate the standard errors of these values. In the table, \textit{NBM}, \textit{VFBM}, \textit{GSFBM}, \textit{VDPM} and \textit{GSDPM} are short for non-Bayesian mixture, finite Bayesian mixture via variational EM, finite Bayesian mixture via Gibbs sampler, DP mixture via variational EM and DP mixture model via Gibbs sampler respectively. For the number of components of the DP mixture via Gibbs sampler, we use the mean of the number of components used in five trials.
\begin{table*}[t]
    \centering
\begin{tabular}{|c|r|l|c|c|c|}
\hline
           & \multicolumn{ 2}{|c|}{Criteria} &         $F^-$ &         $F^+$ &          $\hat{E}$ \\
\hline
\hline
\multirow{ 5}{*}{Syn15} &        NBM &       K=15 &  9.55$\pm$0.63 & {\bf 1.40$\pm$0.24} &  2.12$\pm$0.12 \\
\cline{2-6}
 &       VFBM &       K=15 &  4.11$\pm$0.77 &  2.38$\pm$0.49 &  1.25$\pm$0.13 \\
\cline{2-6}
 &      GSFBM &       K=15 & {\bf 3.19$\pm$0.18} &  2.93$\pm$0.04 & {\bf 1.17$\pm$0.02} \\
\cline{2-6}
 &       VDPM &       K=15 &  3.54$\pm$0.25 &  2.69$\pm$0.48 &  1.18$\pm$0.05 \\
\cline{2-6}
 &      GSDPM &     K=12.6 &  3.84$\pm$0.41 &  2.71$\pm$0.37 &  1.24$\pm$0.03 \\
\hline
\hline
\multirow{ 5}{*}{Syn25} &        NBM &       K=25 &  9.50$\pm$0.58 & {\bf 1.24$\pm$0.21} &  1.94$\pm$0.10 \\
\cline{2-6}
 &       VFBM &       K=25 &  3.73$\pm$1.57 &  2.35$\pm$0.37 &  1.60$\pm$0.20 \\
\cline{2-6}
 &      GSFBM &       K=25 & {\bf 2.63$\pm$0.74} &  2.84$\pm$0.28 & {\bf 0.95$\pm$0.07} \\
\cline{2-6}
 &       VDPM &       K=25 &  3.46$\pm$0.70 &  2.48$\pm$0.48 &  1.06$\pm$0.13 \\
\cline{2-6}
 &      GSDPM &       K=19 &  3.63$\pm$1.13 &  2.71$\pm$0.64 &  1.11$\pm$0.11 \\
\hline
\hline
\multirow{ 5}{*}{Syn50} &        NBM &       K=50 & 10.29$\pm$0.55 & {\bf 0.93$\pm$0.09} &  2.03$\pm$0.10 \\
\cline{2-6}
 &       VFBM &       K=50 &  5.46$\pm$0.65 &  1.26$\pm$0.16 &  1.19$\pm$0.12 \\
\cline{2-6}
 &      GSFBM &       K=50 & {\bf 3.16$\pm$0.32} &  1.60$\pm$0.11 & {\bf 0.85$\pm$0.03} \\
\cline{2-6}
 &       VDPM &       K=50 &  5.14$\pm$0.57 &  1.23$\pm$0.20 &  1.13$\pm$0.07 \\
\cline{2-6}
 &      GSDPM &       K=31 &  5.20$\pm$1.07 &  1.21$\pm$0.17 &  1.13$\pm$0.16 \\
\hline
\hline
\multirow{ 5}{*}{Syn75} &        NBM &       K=75 &  9.59$\pm$0.22 & { 0.71$\pm$0.12} &  1.79$\pm$0.07 \\
\cline{2-6}
&       VFBM &       K=75 &  5.92$\pm$0.86 &  0.70$\pm$0.09 &  1.14$\pm$0.14 \\
\cline{2-6}
 &      GSFBM &       K=75 & {\bf 4.34$\pm$0.60} &  0.81$\pm$0.07 & {\bf 0.89$\pm$0.09} \\
\cline{2-6}
 &       VDPM &       K=75 &  6.04$\pm$0.54 &  {\bf 0.67$\pm$0.08} &  1.14$\pm$0.08 \\
\cline{2-6}
 &      GSDPM &     K=49.6 &  5.76$\pm$0.57 &  0.91$\pm$0.11 &  1.14$\pm$0.08 \\
\hline
\hline
\multirow{ 5}{*}{Syn140} &        NBM &      K=140 & 11.59$\pm$0.32 & {\bf 0.68$\pm$0.06} &  2.31$\pm$0.06 \\
\cline{2-6}
 &       VFBM &      K=140 &  8.49$\pm$0.54 &  1.03$\pm$0.07 &  1.76$\pm$0.12 \\
\cline{2-6}
 &      GSFBM &      K=140 & {\bf 5.43$\pm$0.30} &  1.27$\pm$0.16 & {\bf 1.22$\pm$0.01} \\
\cline{2-6}
 &       VDPM &      K=140 &  8.66$\pm$0.40 &  1.14$\pm$0.21 &  1.80$\pm$0.04 \\
\cline{2-6}
 &      GSDPM &       K=65 &  6.66$\pm$0.26 &  1.45$\pm$0.13 &  1.47$\pm$0.04 \\
\hline
\end{tabular}

    \caption{Test result of synthetic datasets (\%), average of 5 runs}\label{resultSyn15}
\end{table*}

From Table \ref{resultSyn15} we can observe that the average empirical errors of all four Bayesian methods are below 2\%, which means these methods can recover the original model with a relatively small loss of accuracy. Comparing all the methods, GSFBM fits the original model best. Non-Bayesian model gives the worst overall estimation but the best false positive rate. The results of the other approaches are generally comparable. With regards to specific datasets, in the tests on Syn-15 and Syn-25, VDPM is slightly better than GSDPM, and GSDPM is slightly better than VFBM. In the tests of Syn-50, the results of GSDPM and VDPM are very close and both are slightly better than VDPM. In the tests on Syn-75, the three methods give similar results. In the tests on Syn-140, GSDPM outperforms the other two approaches whilst the rest are close.

\begin{figure}[h]
    \centering
    \includegraphics[width=0.35\textwidth, angle=270]{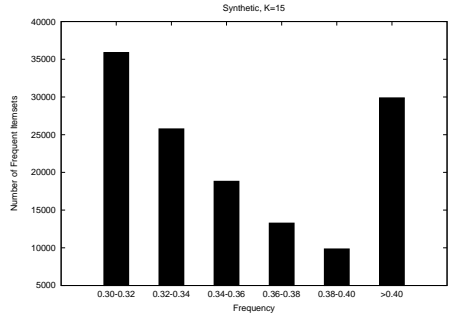}
    \caption{Distribution of the frequent itemsets over frequency of dataset Syn-15}\label{fidist2}
\end{figure}

Although the empirical relative errors of the four approaches are only about 1\%-2\%, the $F^-$ is much higher relatively. This can be explained by the distribution of the frequent itemsets over frequency. Figure \ref{fidist2} is the distribution of frequent itemsets with different frequencies of the synthetic dataset Syn-15. We use this dataset as an example to demonstrate the sensitivity of estimation error on ``edge'' itemsets. The rest of the datasets have similar distributions. From this figure, we can see that there are over 35,000 itemsets in the range of 0.30-0.32, which means about one third of the frequent itemsets are on the ``edge'' of the set of frequent itemsets. Assuming that a model under-estimates each itemset by about 7\%, an itemset with a frequency of 32\% will be estimated as $32\%\times(1-7\%)=29.76\%$, which is below the minimum frequent threshold and the itemset will be labelled as infrequent. This 7\% under-estimation will eventually cause a false negative rate of over 30\%. The reason why the model tends to under-estimate will be discussed later. This example is an extreme circumstance. However, it is clear that with a pyramid like distribution of the frequent itemsets, when we use the $F^-$ and $F^+$ criteria, the actual estimation error will be amplified.

The aim of introducing synthetic datasets is to validate the optimization ability of the five approaches. We want to check whether the algorithms for mixture models can find the right parameters of the wanted model with correct $K$s. The results show that the losses of the five approaches are acceptable with a not-so-large number of components. When the model gets more complicated, the loss caused by the algorithm tends to increase.
\begin{table*}[h]
    \centering
    \subtable{
        \scalebox{0.5}{
\begin{tabular}{|r|l|c|c|c|}
\hline
\multicolumn{ 2}{|c|}{Name} &         \multicolumn{ 3}{|c|}{Chess} \\
\hline
\multicolumn{ 2}{|c|}{Criteria} &         $F^-$ &         $F^+$ &          $\hat{E}$ \\
\hline
\hline
\multirow{ 4}{*}{NBM} &       K=15 & 22.47$\pm$1.24 &  6.68$\pm$0.76 &  3.89$\pm$0.17 \\
\cline{2-5} &       K=25 & 19.71$\pm$1.30 &  7.03$\pm$0.28 &  3.47$\pm$0.19 \\
\cline{2-5} &       K=50 & 17.22$\pm$0.98 &  7.66$\pm$0.77 &  3.12$\pm$0.08 \\
\cline{2-5} &       K=75 & 15.01$\pm$1.05 &  8.62$\pm$0.58 &  2.89$\pm$0.12 \\
\hline
\hline
\multirow{ 4}{*}{VFBM} &       K=15 & 15.10$\pm$0.75 &  9.60$\pm$1.17 &  3.03$\pm$0.10 \\
\cline{2-5} &       K=25 & 13.49$\pm$0.62 &  9.55$\pm$0.27 & {\bf 2.77$\pm$0.08} \\
\cline{2-5} &       K=50 & 14.20$\pm$0.62 &  9.20$\pm$0.91 &  2.83$\pm$0.02 \\
\cline{2-5} &       K=75 & 13.95$\pm$0.91 &   9.4$\pm$0.46 &  2.81$\pm$0.09 \\
\hline
\hline
\multirow{ 4}{*}{GSFBM} &       K=15 & 18.85$\pm$1.94 &  7.03$\pm$0.76 &  3.39$\pm$0.28 \\
\cline{2-5} &       K=25 & 15.63$\pm$1.82 &  8.25$\pm$1.05 &  2.97$\pm$0.17 \\
\cline{2-5} &       K=50 & 14.86$\pm$0.55 &  8.03$\pm$0.42 &  2.83$\pm$0.06 \\
\cline{2-5} &       K=75 &   17.4$\pm$0.6 & {\bf 6.6$\pm$0.39} &  3.07$\pm$0.06 \\
\hline
\hline
\multirow{ 4}{*}{VDPM} &       K=15 & 15.92$\pm$1.67 &  9.56$\pm$1.87 &  3.17$\pm$0.13 \\
\cline{2-5} &       K=25 & 14.31$\pm$0.85 &  8.81$\pm$0.84 &  2.83$\pm$0.05 \\
\cline{2-5} &       K=50 & {\bf 13.49$\pm$0.73} &  9.66$\pm$0.34 &  2.78$\pm$0.10 \\
\cline{2-5} &       K=75 & 14.43$\pm$1.08 &  8.25$\pm$0.77 &  2.77$\pm$0.11 \\
\hline
\hline
     GSDPM &     K=29.6 & 14.46$\pm$0.72 &  8.40$\pm$0.89 &  2.83$\pm$0.05 \\
\hline
\end{tabular}

}
    }
    \quad
    \subtable{
        \scalebox{0.5}{
\begin{tabular}{|r|l|c|c|c|}
\hline
\multicolumn{ 2}{|c|}{Name} &      \multicolumn{ 3}{|c|}{Mushroom} \\
\hline
\multicolumn{ 2}{|c|}{Criteria} &         $F^-$ &         $F^+$ &          $\hat{E}$ \\
\hline
\hline
\multirow{ 4}{*}{NBM} &       K=15 & 71.36$\pm$1.75 &  1.63$\pm$0.12 & 11.59$\pm$0.46 \\
\cline{2-5} &       K=25 & 68.53$\pm$1.12 &  1.40$\pm$0.11 & 10.17$\pm$0.21 \\
\cline{2-5} &       K=50 & 65.72$\pm$1.05 &  1.33$\pm$0.09 &  9.10$\pm$0.10 \\
\cline{2-5} &       K=75 &  25.16$\pm$2.60 &  0.63$\pm$0.01 &  6.81$\pm$0.05 \\
\hline
\hline
\multirow{ 4}{*}{VFBM} &       K=15 & 7.58$\pm$13.17 &  0.75$\pm$0.16 &  5.04$\pm$1.18 \\
\cline{2-5} &       K=25 &  6.27$\pm$7.33 &  {\bf 0.40$\pm$0.05} &  5.70$\pm$0.38 \\
\cline{2-5} &       K=50 &  2.07$\pm$1.80 &  0.59$\pm$0.04 &  5.27$\pm$0.09 \\
\cline{2-5} &       K=75 &  0.97$\pm$0.05 &  0.62$\pm$0.03 &  5.48$\pm$0.01 \\
\hline
\hline
\multirow{ 4}{*}{GSFBM} &       K=15 &  {\bf 0.77$\pm$0.01} &  0.74$\pm$0.06 &  4.09$\pm$0.07 \\
\cline{2-5} &       K=25 &   0.90$\pm$0.02 &  0.72$\pm$0.06 &  4.12$\pm$0.03 \\
\cline{2-5} &       K=50 &   0.90$\pm$0.03 &  0.78$\pm$0.05 &  4.32$\pm$0.05 \\
\cline{2-5} &       K=75 &  0.86$\pm$0.01 &  0.79$\pm$0.01 &  4.29$\pm$0.03 \\
\hline
\hline
\multirow{ 4}{*}{VDPM} &       K=15 &  7.86$\pm$9.08 &  0.83$\pm$0.21 &  4.90$\pm$0.56 \\
\cline{2-5} &       K=25 & 6.09$\pm$10.26 &  0.75$\pm$0.19 &  4.91$\pm$0.66 \\
\cline{2-5} &       K=50 &  1.23$\pm$0.39 &  0.57$\pm$0.10 &  5.44$\pm$0.51 \\
\cline{2-5} &       K=75 &  2.92$\pm$1.36 &  0.63$\pm$0.05 &  5.83$\pm$0.82 \\
\hline
\hline
     GSDPM &     K=18.4 &  { 0.85$\pm$0.01} &   0.70$\pm$0.04 &  {\bf 3.99$\pm$0.03} \\
\hline
\end{tabular}

}
    }
    \subtable{
        \scalebox{0.5}{
\begin{tabular}{|r|l|c|c|c|}
\hline
\multicolumn{ 2}{|c|}{Name} &     \multicolumn{ 3}{|c|}{Accidents} \\
\hline
\multicolumn{ 2}{|c|}{Criteria} &         $F^-$ &         $F^+$ &          $\hat{E}$ \\
\hline
\hline
\multirow{ 4}{*}{NBM} &       K=15 & 21.13$\pm$1.53 &  3.07$\pm$0.25 &  5.05$\pm$0.40 \\
\cline{2-5}
 &       K=25 & 20.60$\pm$1.03 &  2.90$\pm$0.34 &  4.90$\pm$0.24 \\
\cline{2-5}
 &       K=50 & 17.80$\pm$0.52 &  2.76$\pm$0.17 &  4.12$\pm$0.10 \\
\cline{2-5}
 &       K=75 & 13.84$\pm$0.61 & {\bf 2.68$\pm$0.21} &  3.25$\pm$0.13 \\
\hline
\hline
\multirow{ 4}{*}{VFBM} &       K=15 & 13.61$\pm$1.33 &  5.04$\pm$0.22 &  3.69$\pm$0.38 \\
\cline{2-5}
 &       K=25 & 11.78$\pm$1.22 &  4.44$\pm$0.27 &  3.13$\pm$0.23 \\
\cline{2-5}
 &       K=50 & 10.16$\pm$0.60 &  4.07$\pm$0.30 &  2.71$\pm$0.13 \\
\cline{2-5}
 &       K=75 &  9.84$\pm$1.07 &  3.61$\pm$0.26 &  2.63$\pm$0.18 \\
\hline
\hline
\multirow{ 4}{*}{GSFBM} &       K=15 & 13.62$\pm$0.91 &  4.19$\pm$0.40 &  3.58$\pm$0.27 \\
\cline{2-5}
 &       K=25 & 10.85$\pm$0.31 &  4.12$\pm$0.31 &  2.93$\pm$0.11 \\
\cline{2-5}
 &       K=50 &  8.52$\pm$0.91 &  3.78$\pm$0.29 &  2.41$\pm$0.14 \\
\cline{2-5}
 &       K=75 &  8.36$\pm$0.49 &  3.83$\pm$0.21 &  2.37$\pm$0.10 \\
\hline
\hline
\multirow{ 4}{*}{VDPM} &       K=15 & 14.10$\pm$0.47 &  4.82$\pm$0.22 &  3.73$\pm$0.17 \\
\cline{2-5}
 &       K=25 & 11.00$\pm$0.58 &  4.51$\pm$0.27 &  2.95$\pm$0.15 \\
\cline{2-5}
 &       K=50 & 10.64$\pm$0.46 &  3.44$\pm$0.23 &  2.73$\pm$0.08 \\
\cline{2-5}
 &       K=75 & 10.06$\pm$0.72 &  4.00$\pm$0.19 &  2.69$\pm$0.13 \\
\hline
\hline
     GSDPM &    K=114.6 & {\bf 6.80$\pm$0.25} &  3.76$\pm$0.17 & {\bf 2.04$\pm$0.05} \\
\hline
\end{tabular}

}
    }
    \quad
    \subtable{
        \scalebox{0.5}{
\begin{tabular}{|r|l|c|c|c|}
\hline
\multicolumn{ 2}{|c|}{Name} &        \multicolumn{ 3}{|c|}{MS Web} \\
\hline
\multicolumn{ 2}{|c|}{Criteria} &         $F^-$ &         $F^+$ &          $\hat{E}$ \\
\hline
\hline
\multirow{ 4}{*}{NBM} &       K=15 & 47.89$\pm$1.52 &  7.02$\pm$0.57 & 41.77$\pm$1.51 \\
\cline{2-5} &       K=25 & 44.91$\pm$0.67 &  6.44$\pm$0.50 & 38.49$\pm$0.64 \\
\cline{2-5} &       K=50 & 43.96$\pm$0.67 &  5.45$\pm$0.43 & 36.47$\pm$0.50 \\
\cline{2-5} &       K=75 & 39.72$\pm$1.07 &  4.77$\pm$0.32 & 32.17$\pm$1.01 \\
\hline
\hline
\multirow{ 4}{*}{VFBM} &       K=15 & 37.40$\pm$2.41 &  4.52$\pm$0.75 & 30.08$\pm$2.16 \\
\cline{2-5} &       K=25 & 34.11$\pm$0.67 &  3.74$\pm$0.37 & 26.46$\pm$0.52 \\
\cline{2-5} &       K=50 & 32.95$\pm$0.44 &  3.39$\pm$0.11 & 25.17$\pm$0.53 \\
\cline{2-5} &       K=75 & 32.63$\pm$2.54 &  {\bf 3.27$\pm$0.71} & 24.41$\pm$2.51 \\
\hline
\hline
\multirow{ 4}{*}{GSFBM} &       K=15 & 36.53$\pm$0.99 &  4.83$\pm$0.72 & 29.32$\pm$0.51 \\
\cline{2-5} &       K=25 & 33.96$\pm$2.38 &  4.56$\pm$0.59 & 26.67$\pm$1.49 \\
\cline{2-5} &       K=50 & 30.11$\pm$0.72 &  4.14$\pm$0.19 & 23.23$\pm$0.23 \\
\cline{2-5} &       K=75 & 28.49$\pm$0.58 &  4.18$\pm$0.26 & 22.21$\pm$0.59 \\
\hline
\hline
\multirow{ 4}{*}{VDPM} &       K=15 & 38.18$\pm$2.13 &  5.04$\pm$0.68 & 30.22$\pm$1.30 \\
\cline{2-5}&       K=25 & 35.23$\pm$1.37 &  4.06$\pm$0.14 & 26.75$\pm$0.88 \\
\cline{2-5} &       K=50 & 33.37$\pm$0.37 &  3.50$\pm$0.23 & 25.66$\pm$0.37 \\
\cline{2-5} &       K=75 & 33.96$\pm$0.54 &  3.49$\pm$0.39 & 25.51$\pm$0.56 \\
\hline
\hline
     GSDPM &    K=140.6 &    {\bf 28.00$\pm$0.75} &  4.33$\pm$0.25 & {\bf 20.95$\pm$0.43} \\
\hline
\end{tabular}

}
    }
    \caption{The empirical results in percentage (\%) of the four data sets (mean$\pm$std)}\label{result}
\end{table*}
\changes{HE: add comments for each data set}
For the real datasets we tested them with 15, 25, 50 and 75 components respectively. The test results are shown in Figure \ref{result}.

For `chess', Gibbs sampler used 29.6 components on average. Its result
is comparable to the rest of the algorithms with $K=25$, but not as good as VDP with $K=50$ and $K=75$. However, the improvement of VDP when raising truncation level from 25 to 75 is not very great, showing that the proper number of components might be around 30. The false positive and false negative rates look high, but the average estimation error is only around 3\%. This is because 18.04\% of the frequent itemsets' frequencies are just a 2.5\% higher than the threshold, therefore a little under-estimation causes a large false negative rate.

The data set `mushroom' is a quite famous but strange data set. There
are quite a lot of itemsets' which their frequencies are just above
the minimum threshold. Thus a little under-estimation might cause a
great false negative rate. The difference in relative error between
non-Bayesian and Bayesian models is about 6 percent. However the
difference in the false negative rate is large. When training the VDP
model, we find it is very likely that the algorithm is stuck in some local minimum, which causes significant difference among all the 5 trials. In some trials, the $F^-$ are as low as about 1\% while in other trials, the $F^-$ rises to about 25\%. That is the reason that the standard deviations of VDP at truncation level of 15 and 25 are larger than the averages. On the other hand, Gibbs sampler suggests that about 19 components are enough to approximate the distribution of `mushroom'. It gives better results than VDP at truncation level of 50.

In the experiments for `accidents', GS uses 114.6 components on
average, far more than 50. Therefore it gives more accurate results
than the other algorithms. For `MS Web', all three models seriously under-estimate the true probabilities. Yet Bayesian models work better than NBM. GS uses about 140 components to get a result better than finite models. The phenomenon of under-estimation on `MS Web' will be discussed later.

\changes{HE: compare two methods}
Comparing the variational algorithm and Gibbs sampling, a big
advantage of Gibbs sampling is that it is nonparametric, which means
the problem of choosing the number of components can be left to the
algorithm itself. Facing an unknown data set, choosing an appropriate
$K$ is difficult. One has to try several times to determine the
$K$. The tests on the four test cases showed that the Gibbs sampling
can find the proper number. The idea of choosing $K$ automatically is
simply as the following: create a new cluster if no existing cluster
can fit the current data point significantly better than the average
of the whole population. The DP mixture via the Gibbs sampler
implements this idea in a stochastic way. Another advantage is in the
accuracy of Gibbs sampler, as we do not need to make truncation and
decoupling approximations as in variational method. However, a serious
limitation of Gibbs sampler is its speed. As the Gibbs sampler
generates a different number components each time, and due to the different convergence conditions of the methods, we cannot compare the time cost of the two method in a perfect ``fair'' manner.

However, to illustrate the differences in time costs, we show a time
cost analysis of dataset Accidents in Table~\ref{accoff}.
\begin{table}[t]
    \centering
\begin{tabular}{|r|l|r|r|r|}
\hline
\multicolumn{ 2}{|c|}{Name} &     \multicolumn{ 3}{|c|}{Accidents} \\
\hline
\multicolumn{ 2}{|c|}{Criteria} & Iterations &    $T_{Off}$ &  $T_{Off}/I$ \\
\hline
\hline
\multirow{ 4}{*}{NBM} &       K=15 &       15.4 &         132.6 &       8.61 \\
\cline{2-5}
 &       K=25 &       16.8 &       133.4 &       7.94 \\
\cline{2-5}
 &       K=50 &         15.6 &       147.4 &        9.45 \\
\cline{2-5}
 &       K=75 &         14.8 &      145.6 &       9.84 \\
\hline
\hline
\multirow{ 4}{*}{VFBM} &       K=15 &       18.4 &        185.0 &  10.06 \\
\cline{2-5}
 &       K=25 &       18.2 &      187.4 &  12.86 \\
\cline{2-5}
 &       K=50 &         18.0 &      195.4 &  10.86 \\
\cline{2-5}
 &       K=75 &       16.4 &      190.4 &  11.57 \\
\hline
\hline
\multirow{ 4}{*}{GSFBM} &       K=15 &         10.0 &      261.8 &      26.18 \\
\cline{2-5}
 &       K=25 &         10.0 &      276.2 &      27.62 \\
\cline{2-5}
 &       K=50 &       10.2 &      344.4 &  33.75 \\
\cline{2-5}
 &       K=75 &       10.2 &        408.0 &  40.00 \\
\hline
\hline
\multirow{ 4}{*}{VDPM} &       K=15 &         21.0 &      211.2 &  10.06 \\
\cline{2-5}
 &       K=25 &       16.2 &      187.2 &  12.86 \\
\cline{2-5}
 &       K=50 &         17.0 &      184.6 &  10.86 \\
\cline{2-5}
 &       K=75 &       16.4 &      189.8 &  11.57 \\
\hline
\hline
     GSDPM &    K=114.6 &      130.2 &     6812.2 &  51.64 \\
\hline
\end{tabular}

    \caption{Total training time cost and training time per iteration of dataset Accidents (sec), average of 5 runs. $T_{Off}$ is the time used for model training; $T_{Off}/I$ is the training time per iteration}\label{accoff}
\end{table}
The training time costs per iteration of NBM, VFBM and VDPM do not
increase when the $K$ increases. The reason might be due to the
optimized vector computation in Matlab, which makes the increasing of
$K$ less sensitive. The GSFBM and GSDPM involve sampling a multinomial
distribution which cannot be handled as a vector operation in
Matlab. Therefore, their training time cost per iteration is still
relevant to $K$. Generally, NBM is the fastest, variational methods
are a bit slower and sampling methods are the slowest. Although the
training time cost of all methods is $O(NDK)$, NBM does not involve
any complex function evaluation. Variational methods need to calculate
some functions such as logarithm, exponential and digamma
function. The sampling methods need to calculate logarithm and
exponential functions and to generate random numbers. A more time
consuming aspect to sampling is that it needs to update the parameters
after each draw. However, we notice that the number of iterations used by GSFBM is less than that of variational methods. This is because the model of GSFBM is updated after each draw. It can be viewed as an online updating model. On the other hand, variational methods are both updated in the batch mode which normally takes more iterations to converge. The prediction time cost is simpler in comparison with training cost. If the numbers of components of the different models are the same, the prediction time cost should be the same.
\begin{table}[t]
    \centering
    \scalebox{1.0}{
\begin{tabular}{|l|r|}
\hline
           &     $T_{On}$ \\
\hline
     Eclat &     76.31 \\
\hline
      K=15 &      0.37 \\
\hline
      K=25 &   0.46 \\
\hline
      K=50 &    0.66 \\
\hline
      K=75 &   0.79 \\
\hline
   K=114.6 &       0.94 \\
\hline
\end{tabular}

}
    \caption{Itemset generating time cost of dataset Accidents (sec), average of 5 runs. $T_{On}$ is the generating time}\label{accon}
\end{table}

Generally, the Bayesian models take more time than the non-Bayesian model for training. Among the Bayesian models, the variational methods are faster than the two sampling methods. The DP mixture model via Gibbs sampling is the slowest, however the benefit of this approach is that it does not require multiple runs to find the proper $K$.

With the model prepared, the process of generating frequent itemsets by the model is much faster than the Eclat data mining. In most cases the itemset generation process is over 10 times faster than Eclat mining. As the model is irrelevant to the scale of the original dataset and the minimum frequency threshold. The probability models can save more time when we deal with large datasets or we need to mine the dataset multiple times with different thresholds.

\begin{figure*}[t]
    \centering
    \includegraphics[scale=0.40, angle=270]{{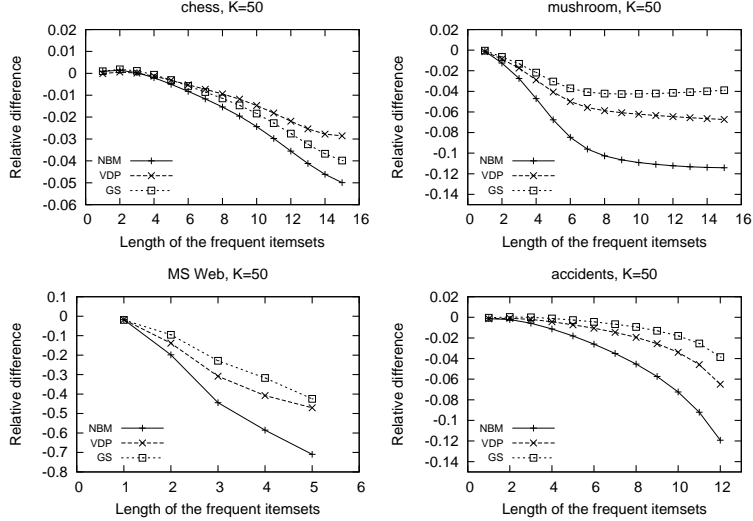}}
    \caption{Relative difference between the models and true frequencies}\label{f4}
\end{figure*}

From the experiment results, we have found several interesting observations about mixture models for frequent itemset discovery. Firstly, as the false negative rates are always much higher than the false positive rates, we observe that the mixture models tend to under-estimate the probabilities of the frequent itemsets. To clarify this, we calculate the empirical errors of the frequent itemsets in a more detailed way. We firstly classify the frequent itemsets into different categories by their lengths. Then we calculate the means of relative difference of all categories. We show the analysis of each data set with 50 components and the Gibbs sampling results in Figure \ref{f4}.

From Figure \ref{f4}, we can see a clear trend that the greater the lengths of the frequent itemsets are, the more the probabilities are under-estimated. The differences of the models' performances are the degree of under-estimation. Similar to the result showed in Table \ref{result}, the degrees of under-estimation of all Bayesian models are better than non-Bayesian mixture.

Another observation is the significant difference of the models' performance between MS Web and the other three data sets. The under-estimation in MS Web is much more serious than the rest. Checking Table \ref{tab_dat}, we notice that MS Web is much sparser than the other three. A further background investigation about the four data sets shows that the difference may be caused by the fact that the correlations between items within the three dense data sets are much stronger than in MS Web, which is sparse. Therefore the distribution of these data records can be better approximated by a mixture model structure. More improvements for the mixture model may be required to achieve a satisfying performance for sparse data sets.

We think the reason for under-estimation is that the mixture model is a mixture of independent models. In independent Bernoulli model, the probabilities of patterns are simply the multiplications of the parameters, which are always under-estimating the correlated item combinations. In mixture models, compensations are made by the assumption of conditional independence. Correlations of the data sets are contained by different components and described by the various group of conditional probabilities. For strongly correlated data sets such as classification data, feature attributes of each class would show high dependencies. These dependencies are strong and simple because the correlated attributes are clustered by the latent classes. Under these circumstances, most correlations are represented by the model thus the under-estimation is tolerable. However, for non-classification data sets where the correlations are not so strong and relatively loose and chaotic, the mixture model cannot hold all the complexity with a feasible number of components. We think this explains why for MS Web data there are severely under-estimation for all three models.

\changes{HE: talk about the advantages of probability models comparing with classic data mining techniques}
Generally, comparing with classic frequent itemset mining, a
well-trained probability model has following benefits. Firstly, the
mixture model can interpret the correlation of the data set and help
people understand the data set while the frequent itemsets is merely a
collection of facts which still need to be interpreted. A probability
model can handle all the kinds of probability queries such as joint
probabilities, marginal probabilities and conditional probabilities
while frequent itemset mining and association rule mining only focus
on high marginal and conditional probabilities. Furthermore,
interesting dependencies between the items, including both positive
and negative, are easier to observe from the model's parameters than
to discriminate interesting itemsets or rules from the whole set of
frequent itemsets or association rules. A second benefit is that generating a set of frequent itemsets is faster than mining the data set if the model is trained. Here we use `chess' as an example since in our case the whole set of frequent itemsets includes 1262028 itemsets thus the mining time is long enough. The average data mining time by Eclat is about 25 seconds while the generation time from the well-trained mixture model takes less than 10 seconds. With the same searching framework, frequent itemset mining obtain the frequency by scanning the database or maintaining a cache of the data set in memory and \textbf{counting} while mixture model generates the probability by various times of multiplications and summation once. At last, the model can serve as a proxy of the entire data set as the model is normally much smaller than the original data set.
\section{Conclusion}
In this paper, we applied finite and infinite Bayesian mixture models via two methods to the frequent itemsets estimation problem. Comparing with earlier non-Bayesian models, Bayesian mixture model can improve estimation accuracy without involving extra model complexity. DP mixture via Gibbs sampler can reach a even better accuracy with proper number of components generated automatically. We tested the Bayesian models and non-Bayesian mixture models on 5 synthetic data sets and 4 benchmark data sets, the experiments showed that in all cases the DP models over performed the non-Bayesian model.


Experiments also showed that all mixture models had the trend of under-estimating the probabilities of frequent itemsets. The average degree of under-estimation increases by the increasing of lengths of the frequent itemsets. For sparse data sets, all mixture models' performances are poor because of the weak correlation between items. Thus one possible avenue for further work would be the use of probability models which explicitly represent sparsity. \changes{Add ref.}\changes{HE: haven't found yet...} We observe that the performance improves as the number of components increases, suggesting some degree of underfitting. Throughout this work we assume that the data was a mixture of transactions of independent models. An alternative approach would be to assume that each transaction is a mixture \cite{Blei:2003:LDA:944919.944937}\changes{Add ref.} or model the indicators distribution as a mixture \cite{Teh04hierarchicaldirichlet}. This might fit the data better.

\section{Appendix}
Here we briefly review the process of EM algorithm for non-Bayesian mixture model.
For all transactions in the data set, if we apply the logarithm, Equation (\ref{prob}) becomes the log-likelihood of the model:
\begin{align}\label{likeall}
    \ln\cL(\cTheta|\T)&=\ln p(\T|\cTheta)\nonumber\\
    &=\sum_{\mu=1}^N\ln\left[\sum_{k=1}^K \pi_k\prod_{i=1}^D \phi_{ik}^{x^\mu_i}(1-\phi_{ik})^{1-x^\mu_i}\right].
\end{align}
However the log-likelihood is hard to optimize because it contains the log of the sum. 
The trick is treating $\Z$ as a random variable. For any distribution $q(\Z)$, the following equation holds:
\begin{align}\label{KL}
    \ln p(\T|\cTheta)=\sum_\Z q(\Z)\ln p(\T|\cTheta)
    =L(q,\cTheta)+\textit{KL}(q||p),
\end{align}
where $\sumz\sim\sum_{z^1=1}^K\dots\sum_{z^\mu=1}^K\dots\sum_{z^N=1}^K$ and
\begin{align}
    L(q,\cTheta)&=\sum_\Z q(\Z)\ln {p(\T,\Z|\cTheta) \over q(\Z)}\label{L(q,theta)}\\
    \textit{KL}(q||p)&=-\sum_\Z q(\Z)\ln {p(\Z|\T,\cTheta) \over q(\Z)}
\end{align}
In Equation (\ref{KL}), $\textit{KL}(q||p)$ is the Kullback-Leibler divergence (KL divergence) between $q(\Z)$ and the true posterior distribution $p(\Z|\T, \cTheta)$. Recall that for any distribution, the KL divergence $\textit{KL}(q||p)\geq0$, with equality if and only if $p(\Z)=p(\Z|\T, \cTheta)$. Therefore based on Equation (\ref{KL}), we have $\ln p(\T|\cTheta)\geq L(q,\cTheta)$. Thus, $L(q,\cTheta)$ can be regarded as a lower bound of the log-likelihood. We can maximize the likelihood by maximizing $L(q,\cTheta)$. 
For $q(\Z)$ we assume it follows a multinomial distribution form:
\begin{align}
    q(\Z)=\prodn q(z^\mu),\text{where }q(z^\mu)\sim \text{Multinomial}(\ctau^\mu),
\end{align}
Thus, we could expand $L(q,\cTheta)$:
\begin{align}
    L(q,\cTheta)=&\sumz q(\Z)\ln p(\T|\Z,\cTheta)+\sumz q(\Z)\ln p(\Z|\cTheta)\eqret
    &-\sumz q(\Z)\ln q(\Z)\label{exL(q,theta)}
\end{align}
All the terms involve standard computations in the exponential family, and the following optimization of the parameters could be solved by a classic multivariate function maximization with constraints. 
\begin{align}
    \taun=&{\pi_k\prodd\phi_{ik}^{x^\mu_i}(1-\phi_{ik})^{1-x^\mu_i} \over \sum_{k'=1}^K\pi_{k'}\prod_{i'=1}^D\phi_{i'k'}^{x^\mu_{i'}}(1-\phi_{i'k'})^{1-x^\mu_{i'}}} \label{tau}\\
    \pi_k=&{1 \over N}\sumn\taun \label{pi}\\
    \phi_{ik}=&{\sumn\taun\xn \over \sumn\taun} \label{phi}
\end{align}
Equation (\ref{pi}) and (\ref{phi}) depend on $\taun$ and Equation (\ref{tau}) depends on $\pi_k$ and $\phi_{ik}$, so the optimizing process alternates between two phases. After the model initialization, first we compute $\taun$ according to Equation (\ref{tau}). This step is called the E-step (Expectation-step). In this step $q(\Z)$ is set to equal $p(\Z|\T, \cTheta^{old})$, causing the lower bound $L(q,\cTheta^{old})$ to increase to the same value as the log-likelihood function $\ln p(\T|\cTheta^{old})$ by vanishing the Kullback-Leibler divergence $\textit{KL}(q||p)$. Then we compute $\pi_k$ and $\phi_{ik}$ according to Equation (\ref{pi}) and (\ref{phi}). This step is called the M-step (Maximization-step). In this step, $q(\Z)$ is fixed and the lower bound $L(q,\cTheta^{old})$ is maximized by altering $\cTheta^{old}$ to $\cTheta^{new}$. As the KL divergence is always non-negative, the log-likelihood function $\ln p(\T|\cTheta)$ increases at least as much as the lower bound does. The EM algorithm iterates the two steps until convergence. A more detailed introduction about EM algorithm is given by \cite{citeulike:873540}.

\bibliographystyle{model1-num-names}
\bibliography{DPM}
\end{document}